\documentclass[sigconf]{acmart}
\usepackage{graphicx}
\usepackage{balance}  % for  \balance command ON LAST PAGE  (only there!)

\usepackage[linesnumbered,ruled,noend]{algorithm2e}
\usepackage{subcaption}
\usepackage{multirow}
\usepackage{booktabs}
\usepackage{color}
\usepackage{hyperref}
\usepackage{appendix}

\AtBeginDocument{%
  \providecommand\BibTeX{{%
    \normalfont B\kern-0.5em{\scshape i\kern-0.25em b}\kern-0.8em\TeX}}}

\copyrightyear{2021}
\acmYear{2021}
\setcopyright{acmcopyright}\acmConference[SIGMOD '21]{Proceedings of the 2021 International Conference on Management of Data}{June 18--27, 2021}{Virtual Event , China}
\acmBooktitle{Proceedings of the 2021 International Conference on Management of Data (SIGMOD '21), June 18--27, 2021, Virtual Event , China}
\acmPrice{15.00}
\acmDOI{10.1145/3448016.3452792}
\acmISBN{978-1-4503-8343-1/21/06}

\SetAlFnt{\small}
\SetAlCapFnt{\small}
\SetAlCapNameFnt{\small}
\SetAlCapHSkip{0pt}
\IncMargin{-\parindent}

\newtheorem{definition}{Definition}

\newcommand{\systems}{Slice~Tuner}
\newcommand{\oneshot}{{\em One-shot}}
\newcommand{\linear}{{\em Conservative}}
\newcommand{\exponential}{{\em Aggressive}}
\newcommand{\moderate}{{\em Moderate}}

\newcommand{\kh}[1]{\textcolor{black}{#1}}

\newcommand{\squishlist}{ 
   \begin{list}{$\bullet$}
    { \setlength{\itemsep}{0pt}      \setlength{\parsep}{3pt} 
      \setlength{\topsep}{3pt}       \setlength{\partopsep}{0pt}
      \setlength{\leftmargin}{1.5em} \setlength{\labelwidth}{1em}
      \setlength{\labelsep}{0.5em} } }

\newcommand{\squishend}{
    \end{list}  } 

\begin{document}

\fancyhead{}
% ****************** TITLE ****************************************

% \title{System T: A Selective Data Acquisition Framework for Accurate and Fair Machine Learning Models}
\title{Slice Tuner: A Selective Data Acquisition Framework for Accurate and Fair Machine Learning Models}

\author{Ki Hyun Tae}
\affiliation{%
%   \institution{KAIST}
%   \institution{Korea Advanced Institute of Science and Technology}
%  \city{Daejeon}
 \country{KAIST}
%  \country{South Korea}
}
\email{kihyun.tae@kaist.ac.kr}

\author{Steven Euijong Whang}
\authornote{Corresponding author}
\affiliation{%
%   \institution{KAIST}
%   \institution{Korea Advanced Institute of Science and Technology}
%  \city{Daejeon}
 \country{KAIST}
%  \country{South Korea}
}
\email{swhang@kaist.ac.kr}

\begin{abstract}
As machine learning becomes democratized in the era of Software 2.0, a serious bottleneck is acquiring enough data to ensure accurate and fair models. Recent techniques including crowdsourcing provide cost-effective ways to gather such data. However, simply acquiring data as much as possible is not necessarily an effective strategy for optimizing accuracy and fairness. For example, if an online app store has enough training data for certain slices of data (say American customers), but not for others, obtaining more American customer data will only bias the model training. Instead, we contend that one needs to {\em selectively acquire data} and propose \systems{}, which acquires possibly-different amounts of data per slice such that the model accuracy and fairness on all slices are optimized. This problem is different than labeling existing data (as in active learning or weak supervision) because the goal is obtaining the right amounts of new data. At its core, \systems{} maintains {\em learning curves} of slices that estimate the model accuracies given more data and uses convex optimization to find the best data acquisition strategy. The key challenges of estimating learning curves are that they may be inaccurate if there is not enough data, and there may be dependencies among slices where acquiring data for one slice influences the learning curves of others. We solve these issues by iteratively and efficiently updating the learning curves as more data is acquired. We evaluate \systems{} on real datasets using crowdsourcing for data acquisition and show that \systems{} significantly outperforms baselines in terms of model accuracy and fairness, even when the learning curves cannot be reliably estimated. 

%We believe \systems{} is a practical tool for suggesting concrete action items based on model analysis.
\end{abstract}

%\keywords{machine learning, selective data acquisition}

\maketitle

\section{Introduction}
\label{sec:introduction}

In the era of Software 2.0~\cite{karpathi}, machine learning (ML) is becoming democratized where successful applications range from recommendation systems to self-driving cars. Software engineering itself is going through a fundamental shift where trained models are the new software, and data becomes a first-class citizen on par with code~\cite{DBLP:journals/sigmod/PolyzotisRWZ18}. Training a model to use in production requires multiple steps including data acquisition and labeling, data analysis and validation, model training, model evaluation, and model serving, which can be a complicated process for ML developers. In response, end-to-end ML platforms~\cite{baylor2017tfx,DBLP:journals/debu/ZahariaCD0HKMNO18} that perform all of these steps have been proposed.

As ML applications become more diverse and possibly narrow, acquiring enough training data is becoming one of the most critical bottlenecks.
We explicitly use the terminology {\em data acquisition} to make a distinction with the broader problem of data collection~\cite{DBLP:journals/corr/abs-1811-03402}, which also includes labeling existing data as in active learning~\cite{DBLP:series/synthesis/2012Settles} or weak supervision~\cite{DBLP:conf/cidr/RatnerHR19}. Instead in data acquisition, the goal is to obtain the right amounts of new data from other data sources. Unlike well-known problems like machine translation where there is decades' worth of parallel corpora to train models, most new applications have little or no training data to start with. For example, a smart factory application for quality control may need labeled images of its own specific products. In response, there has been significant progress in data acquisition research~\cite{DBLP:journals/corr/abs-1811-03402} including dataset discovery~\cite{DBLP:journals/pvldb/NargesianZMPA19}, crowdsourcing~\cite{amazonmechanicalturk}, and even simulator-based data generation~\cite{DBLP:conf/aaai/KimLHS19}. As a result, it is reasonable to assume that new training data can now be acquired at will given enough budget.

\begin{figure}[t]
\begin{subfigure}[t]{0.49\columnwidth}
\centering
\includegraphics[scale=0.55]{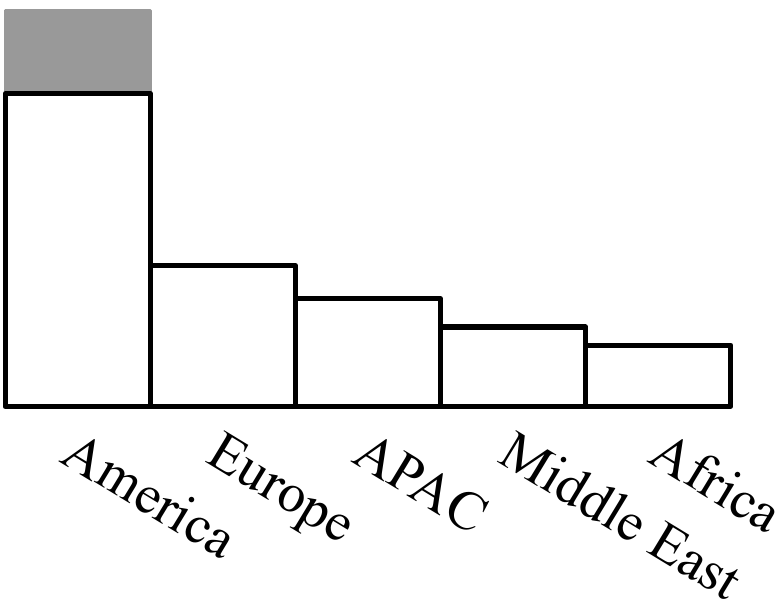}
\vspace{-0.7cm}
\caption{}
\label{fig:customers1}
\end{subfigure}
\begin{subfigure}[t]{0.49\columnwidth}
\includegraphics[scale=0.55]{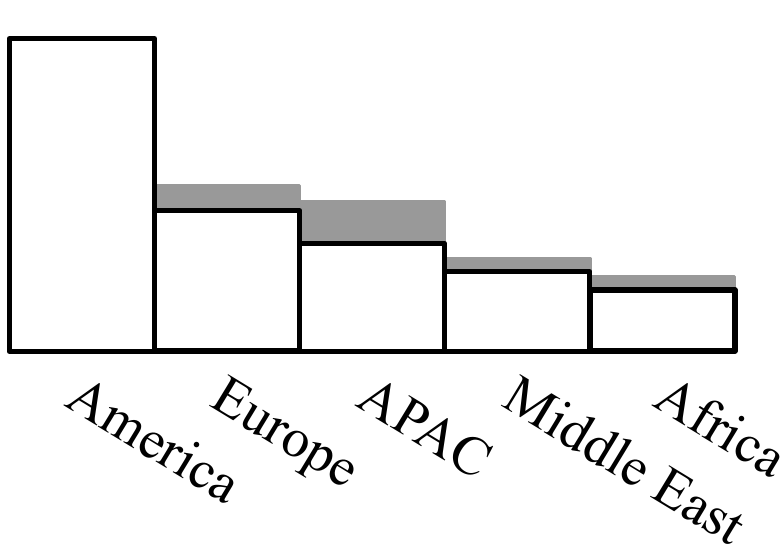}
\vspace{-0.7cm}
\caption{}
\label{fig:customers2}
% \vspace{-0.9cm}
\vspace{-1.2cm}
\end{subfigure}
\caption{Customer slices in different regions where the boxes represent the initial slices (height = slice size) while the gray bars on top indicate the amounts of acquired data per slice. (a) If there is already enough American customer data, acquiring more of it may worsen the data bias. (b) Instead, it is better to acquire possibly-different amounts of data for slices to optimize both the model accuracy and fairness on all slices.}
  \label{fig:customers}
  \vspace{-0.4cm}
\end{figure}

When acquiring data for ML, a common misconception is that more data leads to better models. However, this claim does not always hold when the goal is to improve both the model accuracy and fairness, which are not always aligned. Let us divide the data into subsets called {\em slices}. For example, a company selling apps online may divide its customer purchase data by region as shown in Figure~\ref{fig:customers} (each box indicates a slice where the height is the slice's size). Say that the company not only wants to ensure that its app recommendations are accurate overall, but evenly accurate (i.e., fair) for customers in different regions. Since the company already has enough data for American customers (the slice size is largest), acquiring more American customer data (gray part in Figure~\ref{fig:customers1}) is not only unnecessary, but may bias the training data and influence the model's accuracy on other regions. 

We thus propose a {\em selective data acquisition} framework called \systems{}, which determines how much data to acquire for each slice. Not all slices have the same data acquisition cost-benefits, which means that acquiring the same amounts of data may result in different improvements in model loss for different slices. Hence, the na\"ive strategy of acquiring equal amounts of data per slice is not necessarily optimal. Alternatively, acquiring data such that all slices end up having equal amounts of training data is not always optimal either. (This approach is similar to the Water filling algorithm~\cite{Proakis2007}; details in Section~\ref{sec:selectivedatacollection}.) Instead, we would like a data acquisition strategy such that the models are accurate and similarly-accurate for different slices. To measure model accuracy, we use loss functions like logistic loss. For fairness we use an extension of equal error rates~\cite{DBLP:conf/pods/Venkatasubramanian19,DBLP:conf/www/ZafarVGG17}, which is one of the widely-accepted definitions of fairness along with other traditional ones like demographic parity~\cite{DBLP:conf/kdd/FeldmanFMSV15} and equalized odds~\cite{DBLP:conf/nips/HardtPNS16}. Interestingly, equal error rates is a familiar concept in systems research where it is important to have similar performance across different partitions of data. The key challenge is to figure out the different cost benefits of the slices and estimate how much data to acquire for each slice given a budget. In Figure~\ref{fig:customers2}, we may want to acquire even amounts of data only for non-America regions.

% where we take the average difference between an ``underperforming'' slice versus the rest of the data (see the exact definition in Section~\ref{sec:preliminaries}).

\begin{figure}[t]
  \includegraphics[width=\columnwidth]{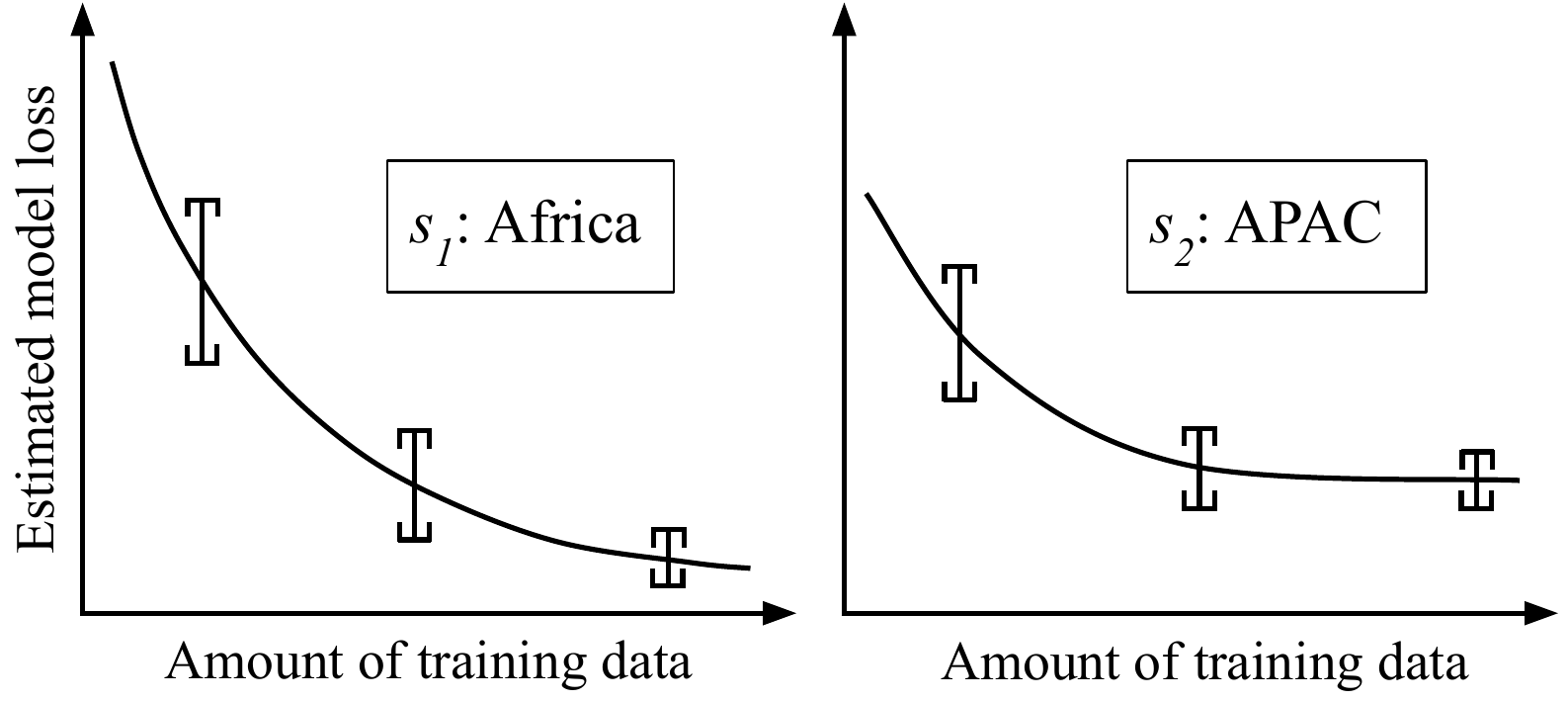}
  \vspace{-0.4cm}
  \caption{Hypothetical learning curves of two slices.} 
  % Each learning curve can be estimated using samples of how the model loss decreases as more data is acquired plus the assumption that the curves follow a power law. }
  \vspace{-0.4cm}
  \label{fig:learningcurve}
%   \vspace{-0.1cm}
\end{figure}

At the core of \systems{} is the ability to estimate the {\em learning curves} of slices, which reflect the cost benefits of data acquisition. It is well known that the impact of data acquisition on model loss is initially large, but eventually plateaus~\cite{DBLP:conf/kdd/ShengPI08}. That is, lowering the loss becomes difficult to the point where it is not worth the data acquisition effort. Figure~\ref{fig:learningcurve} shows hypothetical learning curves of two slices. Recently, multiple studies~\cite{DBLP:conf/nips/ChenJS18,baidu2017deep,DBLP:conf/ijcai/DomhanSH15} show that these curves usually follow a power law according to empirical results in machine translation, language modeling, image classification, and speech recognition. Given the learning curves, \systems{} ``tunes'' the slices by determining how much data to acquire for each slice such that the model accuracy and fairness on all slices are optimized while using a limited budget for data acquisition.  

% \begin{subfigure}[t]{0.49\columnwidth}
% \includegraphics[scale=0.55]{customers2.pdf}

As a toy example, suppose there are two slices $s_1$ and $s_2$ of the same size. Say the current model losses of the two slices are 5 and 3, respectively. Hence, the loss of the entire data is 4. According to Definition~\ref{def:unfairness} in Section~\ref{sec:preliminaries}, we compute the unfairness as avg$\left\{|5 - 4|, |3 - 4|\right\} = 1$. Now suppose we estimate their two learning curves as shown in Figure~\ref{fig:learningcurve}. Notice that the curve of $s_2$ is rather flat, so acquiring more data does not have as much benefit as acquiring for $s_1$. Suppose we have a budget of acquiring examples using crowdsourcing. Since $s_1$ has a better cost-benefit, we may decide to only acquire examples for $s_1$. The actual numbers would depend on the result of the optimization problem. As a result, say the model losses of $s_1$ and $s_2$ are now 2 and 3, respectively, and that the entire data now has a loss of 2.4 ($s_1$ is now larger than $s_2$, so the average loss is closer to 2 than 3). In that case, the unfairness improves by decreasing to avg$\left\{|2 - 2.4|, |3 - 2.4|\right\} = 0.5$.

While learning curves are potentially useful, in reality we cannot assume they will be generated in perfection. Instead, the key technical challenge we address is generating learning curves that are ``reliable enough'' to still benefit \systems{} using data management techniques. There are largely two obstacles. First, slices may not have enough data for the model losses to be precisely measured. Second, acquiring data for one slice may ``influence'' the model's losses on other slices and thus distort their learning curves. \systems{} solves these problems by iteratively updating the learning curves as more data is acquired so that the learning curves are up-to-date. In addition, \systems{} exploits the learning curves by doing a relative comparison of losses, so perfectly predicting the absolute losses in each learning curve is not necessary. 

In our experiments, we use various real datasets to show that \systems{} outperforms other baselines in terms of model loss and unfairness, even when the learning curves are not reliable. We also demonstrate the real scenario of acquiring new data using crowdsourcing via Amazon Mechanical Turk~\cite{amazonmechanicalturk} for the image dataset UTKFace~\cite{zhifei2017cvpr} and show that \systems{} is effective even if the data is acquired from a completely different data source. 

A remaining question is how to define the slices themselves. Data slicing can either be done manually~\cite{tfma} or automatically~\cite{slicefinder}. While \systems{} can run on any set of slices, a desirable property of a slice is to be unbiased such that acquiring any example that belongs to it has a similar effect on model loss as any other possible example. In this work, we assume the slices are provided by the user, but also discuss possible solutions in Appendix ~\ref{sec:slicingmethods}.
% our technical report~\cite{slicetunertr}.
% Appendix ~\ref{sec:slicingmethods}.

To democratize ML, it is important to help users to not only analyze their models, but also fix any problems easily. To our knowledge, \systems{} is the first system to provide concrete action items of how much data to acquire per slice to make models both accurate and fair. We also release our code and crowdsourced dataset as a community resource~\cite{github}.

The rest of the paper is organized as follows:
\squishlist
%    \item We present the related work (Section~\ref{sec:relatedwork}).
    \item We cover preliminaries and formulate the problem of selective data acquisition (Section~\ref{sec:problemdefinition}).
    \item We describe \systems{}'s architecture (Section~\ref{sec:architecture}).
    \item We propose accurate and efficient methods for estimating the learning curves of slices (Section~\ref{sec:learningcurve}).
    \item We propose selective data acquisition algorithms (Section~\ref{sec:algorithms}).
%    \squishlist
%        \item \oneshot{}: Assumes that slices are independent and suggests how much data to acquire in one step.
%        \item {\em Iterative}: Repeatedly updates the learning curves as more data is acquired and invokes \oneshot{}.
%    \squishend
    %\item We discuss methods for finding slices for data acquisition (Section~\ref{sec:slicingmethods}).
    \item We evaluate \systems{} on real datasets and show that it outperforms baselines by obtaining better model accuracy and fairness results using the same data acquisition budget, even if the learning curves cannot be reliably estimated (Section~\ref{sec:experiments}).
\squishend

\section{Problem Definition}
\label{sec:problemdefinition}

%We define data slices, model the effort for data acquisition, define measures for accuracy and fairness, and define the selective data acquisition problem.

\subsection{Preliminaries}
\label{sec:preliminaries}

\paragraph*{Data Slicing}

We denote by $D$ the training data set with examples where each example $e \in D$ has features and a label.
Without loss of generality, we assume all examples have equal weights when training a model.
$D$ can be divided into slices $S = \{s_i\}_{i=1}^n$ where each $s_i \subseteq D$. We assume that the slices partition $D$, i.e., $\bigcup_i s_i = D$. A typical way to define a slice is to use conjunctions of feature-value pairs, e.g., $region = Europe \wedge gender = Female$. We can also use the label feature for the slicing. For example, the Fashion-MNIST dataset~\cite{xiao2017/online} contains images that represent items, e.g., shoes and shirts. Here, we can define a slice for each item. The slices can be found automatically or set manually by a domain expert. Although there is no restriction in defining a slice, a desirable property is that the slice is unbiased such that adding a new example to it has a similar effect on the model accuracy as any other example. We assume the slices are provided by users, but also discuss possible solutions in Appendix ~\ref{sec:slicingmethods} where the idea is to iteratively divide slices until they are unbiased using decision tree methods.

% our technical report~\cite{slicetunertr} where the idea is to iteratively divide slices until they are unbiased using decision tree methods.

% Appendix ~\ref{sec:slicingmethods} where the idea is to iteratively divide slices until they are unbiased using decision tree methods.

% We also define the {\em complement} of a slice $s$ to be the other examples in the entire dataset, i.e., $D - s$.
% For example, a slice that contains all regions of Figure~\ref{fig:customers1} is not desirable because there is a bias towards American customers. That is, adding an American customer example is not as helpful as say adding a European customer example. In this case, it is better to divide the slice until the smaller slices no longer have the bias. On the other hand, we do not want a slice to be too small (e.g., a handful of customers) in the sense that acquiring data has negligible impact on the overall model accuracy. We discuss strategies for finding slices without bias in Section~\ref{sec:slicingmethods}. 

\paragraph*{Model Accuracy and Fairness Measures}

We assume a model $M$ is trained on $D$ or its subsets. We also assume a classification loss function $\psi(s, M)$ that returns a performance score on how well $M$ predicts the labels of the dataset $s$. A common loss function for binary classification is {\em log loss}, which is $\frac{1}{m} \sum_{i=1}^m -y^{(i)} \log \hat{y}^{(i)} - (1 - y^{(i)})\log (1 - \hat{y}^{(i)})$.
% defined using cross entropy:
% ($\frac{1}{m} \sum_{i=1}^m -y^{(i)} \log \hat{y}^{(i)} - (1 - y^{(i)})\log (1 - \hat{y}^{(i)})$).
% \begin{align*}
%     LogLoss = \frac{1}{m} \sum_{i=1}^m -y^{(i)} \log \hat{y}^{(i)} - (1 - y^{(i)})\log (1 - \hat{y}^{(i)}).
% \end{align*}
% A perfect classifier $M$ has a log loss of zero, and a random classifier $-ln(0.5)=0.693$. 
Our setting can be generalized to other ML problems (multi-class classification and regression) by using the appropriate loss functions. Throughout the paper, we will use loss to measure accuracy.

Our notion of fairness extends equalized error rates~\cite{DBLP:conf/pods/Venkatasubramanian19}, which has the definition $Pr(\hat{y} \neq y | z = 0) \approx Pr(\hat{y} \neq y | z = 1)$ where $\hat{y}$ is the model prediction, $y$ is the label, and $z$ is a sensitive attribute like gender. We make the straightforward extension that the model error rates must be similar among any number of groups that are not necessarily defined with sensitive attributes. Equalized error rates is one of the standard fairness measures along with demographic parity~\cite{DBLP:conf/kdd/FeldmanFMSV15} and equalized odds~\cite{DBLP:conf/nips/HardtPNS16}. We choose equalized error rates because it is important for any ML product that needs to provide similar service quality to customers of different demographics. This fairness is also a familiar notion in systems research where a system should perform well on different partitions of data.

%\kh{While most existing fairness research typically assumes two sensitive groups based on a sensitive attribute (e.g., $gender = male$, $gender = female$), we extend the notion to any number of slices as ~\cite{pmlr-v80-agarwal18a} which defines the unfairness of demographic parity and equalized odds for multiple groups.}

% Suppose we have a model $M$. We say a slice $s$ is {\em underperforming} if it has a higher loss than its complement, i.e., $\psi(s, M) > \psi(D - s, M)$. Let us denote the set of underperforming slices as $S_u \subseteq S$.

We propose the following unfairness measure that is used to obtain equalized error rates:
\begin{definition}
\label{def:unfairness}
The {\em unfairness} of the slices is defined as the average absolute difference between the loss of a slice and the entire data $D$: 
% $\mathrm{avg}_{s_i \in S} |\psi(s_i,M) - \psi(D,M)|$
\[\mathrm{avg}_{s_i \in S} |\psi(s_i,M) - \psi(D,M)|.\]
\end{definition}
Notice that the unfairness measure does not require sensitive attributes and can thus be used on any dataset. We can also think of other variations such as computing the maximum absolute difference instead of the average.

% \begin{definition}
% \label{def:unfairness}
% The {\em unfairness} of the slices is defined as the average difference between the loss of an underperforming slice and its complement:
% % \[\mathrm{avg}_{s \in S_u} \psi(s,M) - \psi(D - s,M).\]

% \end{definition}
% Notice that the unfairness measure does not require sensitive attributes and can thus be used on any dataset. We can also think of other variations such as computing the maximum difference instead of the average.

%As we shall see later, in addition to ensuring model fairness, enforcing disparate mistreatment also has the side effect of preventing a slice from having too much data and obtaining an unusually-high accuracy, which may negatively affect the accuracy of other slices.

\paragraph*{Data Acquisition Cost}

We assume that any data acquisition technique can be used to acquire data by paying a certain cost. Nowadays, dataset discovery systems~\cite{DBLP:journals/pvldb/NargesianZMPA19} can help find relevant datasets. In addition, crowdsourcing tools like Amazon Mechanical Turk~\cite{amazonmechanicalturk} or real-world simulators~\cite{DBLP:conf/aaai/KimLHS19} can be used to generate data at will.

We abstract the data acquisition methods and define a cost function $C(s)$ that returns the cost to acquire an example in slice $s$. Even for the same data source, the data acquisition cost may vary by slice. For example, acquiring face images of large groups (e.g., $race = White$) may be easier than images of smaller groups (e.g., $race = Native\ Hawaiian$). We assume that within the same slice, the cost to acquire an example is the same. As more examples are acquired for $s$, $C(s)$ may increase possibly because data becomes scarcer. However, we assume that data is acquired in batches (e.g., when using Amazon Mechanical Turk, we use a fixed budget at a time) and that $C(s)$ is a constant for each batch.

\paragraph*{Slice Dependencies}

Slices may have {\em dependencies} where acquiring data for one slice may influence the learning curves of other slices. For example, if there are two independent slices $s_1$ and $s_2$, and we acquire too much data for $s_1$, the model may overfit and have a worse accuracy on $s_2$. If $s_2$ is similar to $s_1$ content-wise, then the accuracy on $s_2$ can actually increase as well. If there are no dependencies, the slices are considered {\em independent} of each other. We discuss how to handle slice dependencies in Section~\ref{sec:dependentslices}.

\subsection{Selective Data Acquisition}
\label{sec:selectivedatacollection}

We now define the problem of selective data acquisition.

\begin{definition}
\label{def:selectivedatacollection}
Given a set of examples $D$, its slices $S = \{s_i\}_{i=1}^n$, a trained model $M$, a data acquisition cost function $C$, and a data acquisition budget $B$, the selective data acquisition problem is to acquire $d_i$ examples for each slice $s_i \in S$ such that the following are all satisfied:
\squishlist
    \item The average loss $\psi(D,M)$ is minimized,
    \item The unfairness $\mathrm{avg}_{s_i \in S} |\psi(s_i,M) - \psi(D,M)|$ is minimized, and
    \item The total data acquisition cost $\sum_i C(s_i) \times d_i = B$.
\squishend
\end{definition}
We note that minimizing loss and unfairness are correlated, but not necessarily the same and thus need to be balanced (Section~\ref{sec:balancinglambda} studies the tradeoffs). In some cases, making sure slices have similar losses may also result in the lowest loss. For example, if there are two independent slices with identical learning curves, and one of them has less data, then simply making the two slices have the same amount of data results in the optimal solution. However, there are cases where the two objectives are not aligned. Continuing our example, suppose that the two slices now have different learning curves where the slice with less data has a curve that is lower than the other curve and also decreases more rapidly. In this case, acquiring data for the smaller slice would lower loss, but increase unfairness. Instead, the optimal solution could be to also acquire some data for the larger slice to lower the loss without sacrificing too much fairness.

\kh{Our problem can be viewed as a special type of multi-armed bandit problem as we discuss in Section~\ref{sec:relatedwork}.}

%\kh{One may think that our problem is related to a multi-armed bandit problem~\cite{10.1023/A:1013689704352} (especially, rotting bandits~\cite{rottingbandit}) where each slice can be viewed as an arm and the reward of pulling an arm can be the model loss improvement via data acquisition. We discuss the details in Section~\ref{sec:relatedwork}.}

\begin{figure}[t]
\begin{subfigure}[t]{0.49\columnwidth}
\centering
\includegraphics[scale=0.55]{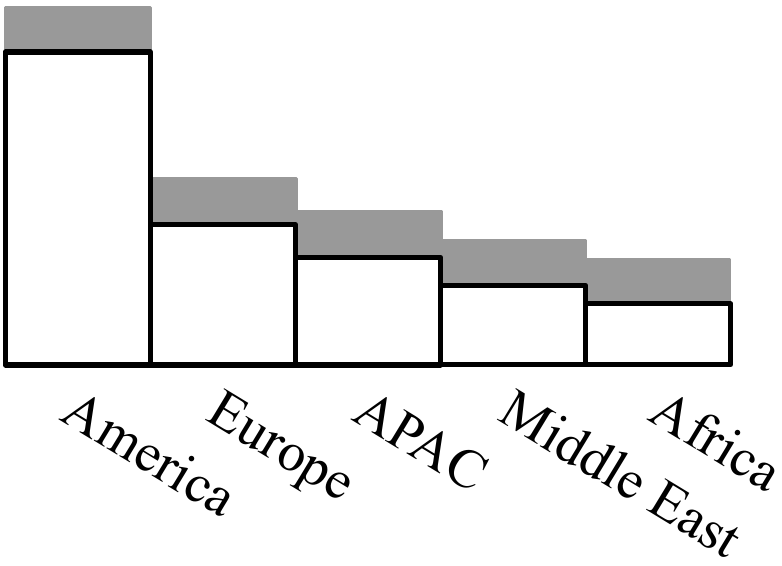}
\vspace{-0.6cm}
\caption{}
\label{fig:baseline1}
\end{subfigure}
\begin{subfigure}[t]{0.49\columnwidth}
\includegraphics[scale=0.55]{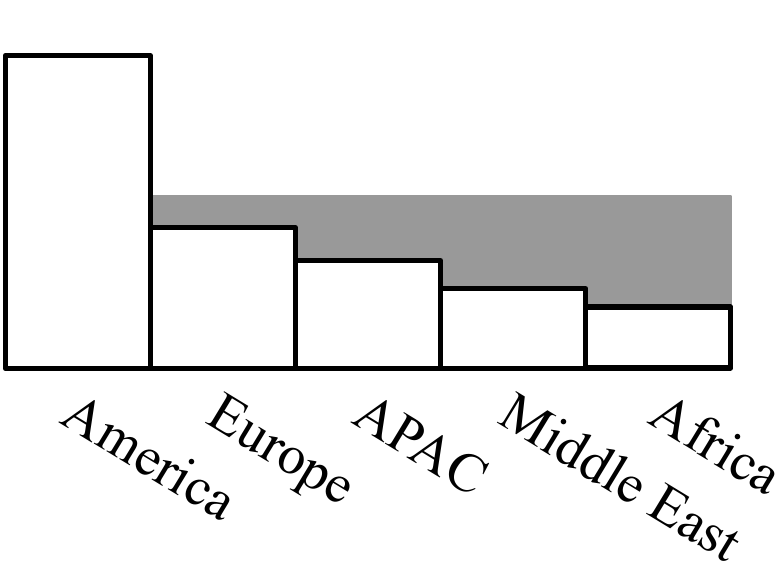}
\vspace{-0.6cm}
\caption{}
\label{fig:baseline2}
\end{subfigure}
\vspace{-0.2cm}
\caption{The two baselines for data acquisition: (a) acquire similar amounts of data per slice, (b) acquire data such that the slices have similar sizes in the end (Water filling alg.).}
  \label{fig:baseline}
  \vspace{-0.5cm}
%   \vspace{-0.4cm}
\end{figure}

\paragraph*{Baselines for Comparison}
Recall we described two data acquisition baselines in Section~\ref{sec:introduction}. Both methods are reasonable starting points, but do not optimally solve our problem in Definition~\ref{def:selectivedatacollection}. Suppose there are two independent slices $s_1$ and $s_2$. The first baseline is to acquire equal amounts of data for all slices (Figure~\ref{fig:baseline1}). This approach does not perform well if the two slices have significantly different learning curves. In a worst-case scenario, $s_1$ may already have a low loss and does not need more data acquisition whereas $s_2$ may have a high loss and can benefit from more data. In this case, acquiring equal amounts of data will result in both suboptimal loss and unfairness. The second baseline is to acquire data such that all slices have similar amounts of data in the end, which can be viewed as a Water filling algorithm (Figure~\ref{fig:baseline2}). The implicit assumption is that all slices require similar amounts of data to obtain similar losses. However, this assumption does not hold if the two slices have different losses even if they have the same size. Continuing from the above worst-case scenario, if $s_1$ is smaller than $s_2$, then we will end up acquiring data for $s_1$ unnecessarily and again get suboptimal results. What we need instead is a way to utilize the learning curves and solve an optimization problem to determine how much data to acquire for each slice. 

Another possible baseline is to acquire data in proportion to the original data distribution~\cite{DBLP:conf/nips/ChenJS18}. However, this approach does not fix data bias at all, so we considered it strictly worse than the baselines above. Also, we re-emphasize that active learning is not a comparable technique because it labels existing data instead of acquiring new data. 

\section{System Overview}
\label{sec:architecture}

\begin{figure}[t]
  \includegraphics[width=0.95\columnwidth]{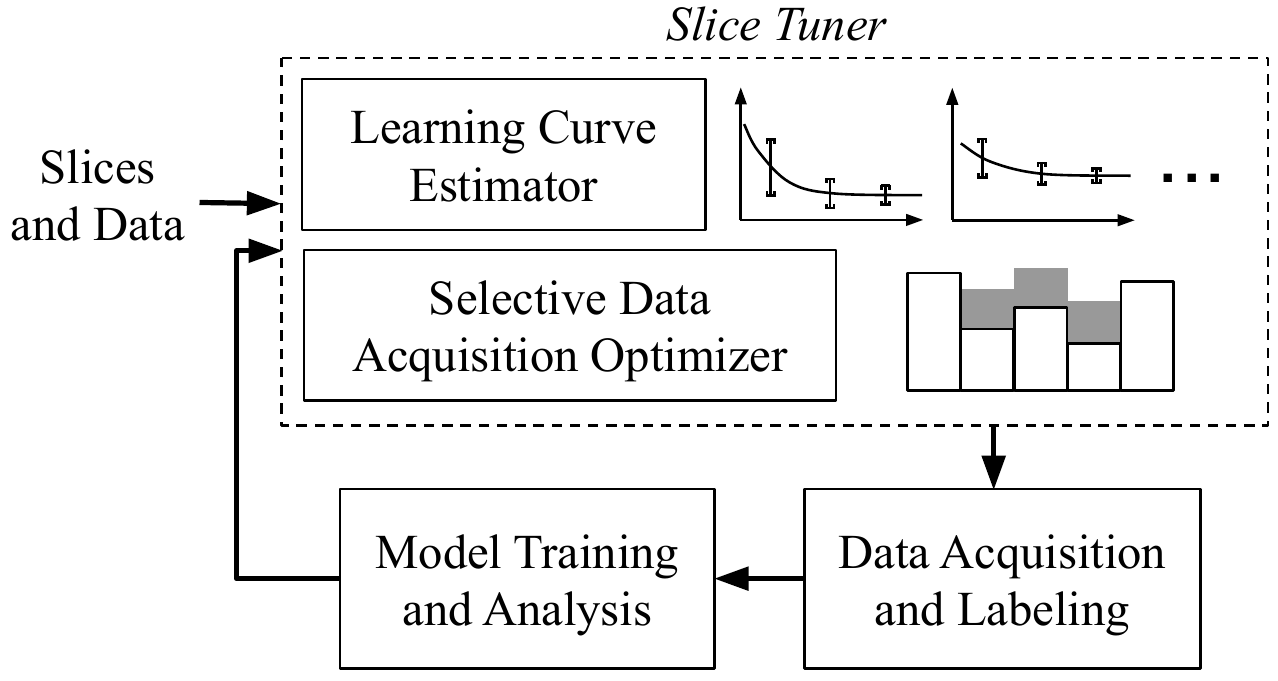}
  \vspace{-0.1cm}
  \caption{The input of \systems{} are the slices with their data. \systems{} consists of two main components: the Learning Curve Estimator and Selective Data Acquisition Optimizer. The learning curves may be iteratively updated as more data is acquired by the Data Acquisition component. \systems{} suggests how much data to acquire per slice.}
  \label{fig:systems}
  \vspace{-0.5cm}
\end{figure}

We describe the overall workflow of \systems{} as shown in Figure~\ref{fig:systems}. \systems{} receives as input a set of slices and their data and estimates the learning curves of the slices by training models on samples of data. We explain the details of the estimation methods in Section~\ref{sec:learningcurve}. Next, \systems{} performs the selective data acquisition optimization where it determines how much data should be acquired per slice in order to minimize the loss and unfairness. As data is acquired, the learning curves can be iteratively updated. We propose selective data acquisition algorithms in Section~\ref{sec:algorithms}. 

We discuss the runtime requirements of \systems{}. Looking at Figure~\ref{fig:systems}, we consider the Data Acquisition step to be the most expensive process done as a batch process, especially if manual crowdsourcing is used. Even if there are tools for dataset searching and simulation, there is still a fair amount of manual work involved. As a result, it is critical for \systems{} to minimize the amount of data acquisition at the expense of possibly using more computation for estimating learning curves and performing the optimization.

\section{Learning Curve}
\label{sec:learningcurve}

A learning curve is a projection of how a model trained on the entire dataset will perform on a particular slice $s$ as a function of the number of examples in $s$. Assuming that the examples are helpful to the model training, we expect the loss to decrease as more examples are added. However, this trend may not always hold due to multiple factors: the examples may be noisy and actually harm the model training, the examples may be biased and only represent a small part of the slice, or the model training itself may be unstable, all of which may result in non-monotonic behavior. Despite the complexity, we believe it is reasonable to assume that more training data on an unbiased slice generally leads to lower loss, but that the benefits have diminishing returns. Unlike existing work, a significant challenge for \systems{} is to plot the learning curves on slices, which can be arbitrarily smaller than the entire data. We first discuss how to efficiently estimate learning curves using data management in this section and then how to handle unreliable learning curves in Section~\ref{sec:algorithms}.

\subsection{Estimation}
\label{sec:estimation}

\begin{figure}[t]
  \includegraphics[width=1\columnwidth]{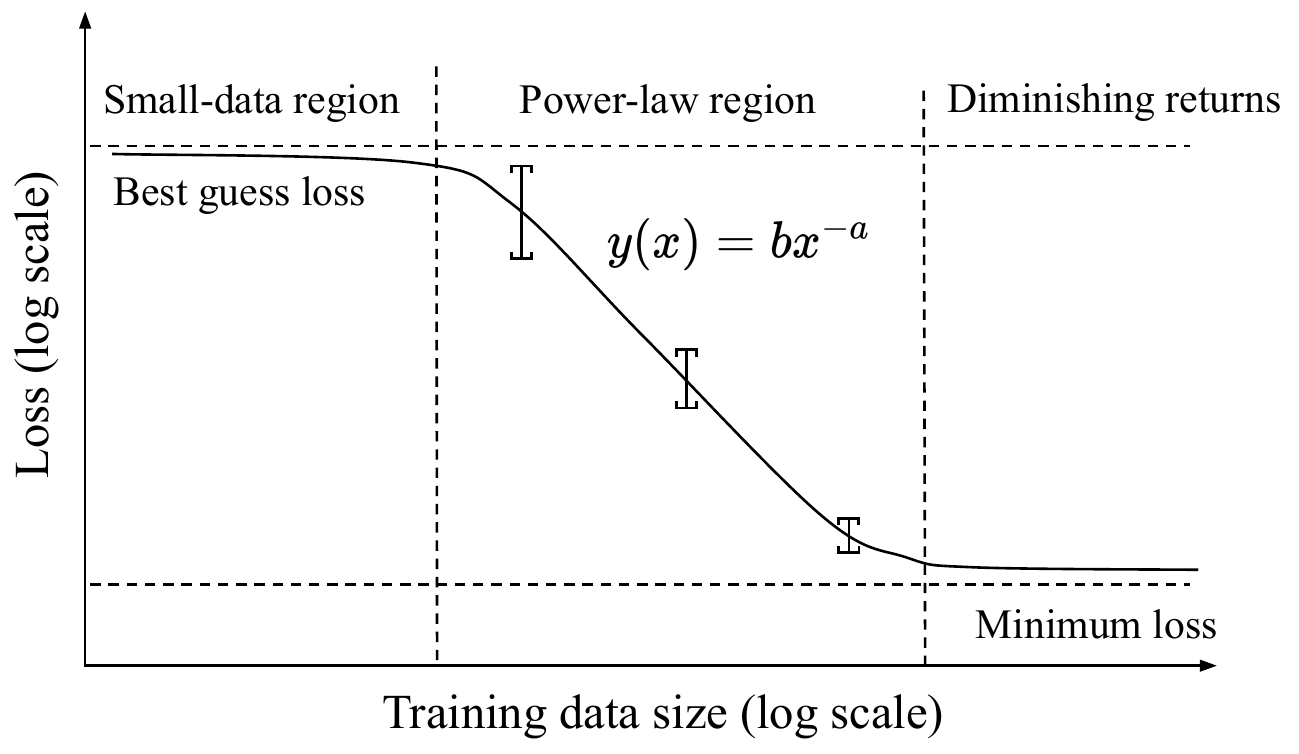}
%   \vspace{-0.4cm}
  \caption{A detailed view of how loss changes against more training data based on work by Baidu \protect\cite{baidu2017deep}. There are three regions: (1) {\em small-data region}: the slice is too small to the extent that the model predictions are more or less random, (2) {\em power-law region}: the curve is of the form $y = bx^{-a}$, and (3) {\em diminishing-returns region}: there is enough data, so acquiring more does not help reduce the loss.}
  \label{fig:curveanalysis}
 \vspace{-0.5cm}
\end{figure}

A key property we exploit is that training data benefits the model accuracy, but more collection has diminishing returns. A recent work from Baidu~\cite{baidu2017deep} conducted an analysis on learning curves (see Figure~\ref{fig:curveanalysis}). The curve starts with the {\em small-data region}, where models try to learn from a small number of training data and can only make ``best guess'' predictions. According to our experience, tens of examples is enough to move beyond this region. For more data, we can see a {\em power-law region} of the form $y = bx^{-a}$, where new training examples provide useful information to improve the predictions. For real-world applications, a lower bound error may exist due to errors such as mislabeled data that cause incomplete generalization. Hence, we then see a {\em diminishing-returns region} where there is a minimum loss that cannot be reduced. In this case, one may also model the learning curve using the form $y = bx^{-a} + c$~\cite{baidu2017deep} where $c$ is the lower-bound loss. However, if not enough data has been acquired to observe diminishing returns, fitting the curve $y = bx^{-a}$ works better~\cite{baidu2017deep}. Another study~\cite{DBLP:conf/ijcai/DomhanSH15} compares 11 parametric models including variations of exponential models and custom models. Based on these works, a power-law curve fits as well as any other curve.

To fit a learning curve of a slice, we first divide its data into train and validation sets. Although the train set may be small, we do assume a validation set that is large enough to evaluate models. This assumption is reasonable in the common setting where we start with some initial data and would like to acquire more examples. We then train models on random subsets of the train set with different sizes and generate data points by evaluating the models on the validation set. How many data points to generate depends on how much time we are willing to invest to estimate the learning curve. Each time we train a model on a subset of data, we also combine that data with the rest of the slices. (Section~\ref{sec:efficientimplementation} presents a more efficient method for generating data points.) Since the model losses on smaller subsets are less reliable (depicted as having high variance Figure~\ref{fig:curveanalysis}), \kh{we give weights to subsets that are proportional to their sizes} and use a non-linear least squares method~\cite{DBLP:journals/midm/FigueroaZKN12} for curve fitting. We further improve reliability by drawing multiple curves (we use 5) and averaging them at the expense of more computation.

%\kh{We thus give more weight to larger subsets by the size of the data using a non-linear least squares method~\cite{DBLP:journals/midm/FigueroaZKN12}. We further improve the learning curve accuracy by drawing multiple curves and averaging them at the expense of more computation.}

In the worst case when all subsets are in the small-data region, then \systems{} may not benefit from learning curves, but in that case would fall back to performing like baselines. If there is enough data in a slice to be in the diminishing return region, we do not need to do any special handling because \systems{} will simply acquire data for other slices that need more data.

\subsection{Efficient Implementation}
\label{sec:efficientimplementation}

We now discuss better data management techniques for more efficient learning curve fitting. Given the slices $S$, suppose we generate for each slice $K$ random subsets of data to fit a power-law curve. In addition, we may repeatedly update the learning curves $U$ times using our iterative algorithms in Section~\ref{sec:dependentslices}. We would thus have to train a model $|S| \times K \times U$ times. A typical setting in our experiments is $|S| = 10$, $K = 10$, and $U = 5$, which means that we may train a model 500 times. Moreover, if we train a model on a subset of a slice plus the rest of the slices, then the rest of the slices could be relatively too large for the model to properly train on the subset due to the data bias. Moreover, each model training may take a long time because the rest of the slices are repeatedly used for training in their entirety.

We thus propose an amortization technique to drastically reduce the number of model trainings. Instead of taking an X\% subset of one slice and leaving the rest as is to train each model, we take X\% subsets of all slices and train a model. This model is then evaluated on each of the slices to generate different learning curves independently. This approach assumes that taking subsets of other slices does not affect the model accuracy on the current slice. In reality, this independence assumption does not hold, so \systems{} solves the problem by periodically updating the learning curves as we explain in Section~\ref{sec:dependentslices}. As a result, the number of model trainings becomes independent of the number of slices and reduces to $O(K \times U)$. In our example above, number 500 reduces to about 50. In addition, each model training is on average faster than the exhaustive approach above because it is performed on smaller data (X\% subsets). Finally, the learning curves can be generated in parallel, so using more nodes further reduces the wall-clock time. 
%We verify these results in Section~\ref{sec:efficientlearningcurvegeneration}.

Another possible optimization is to use a chain of transfer learning where a model trained on one subset is re-used to train the model of the next subset faster and so on. This approach has the potential to reduce the average model training time, although the time complexity remains the same. However, a downside is that hyperparameter tuning becomes significantly more complicated where we may need to use different learning rates for subsequently-trained models, so we do not use this optimization in our experiments.

% First, instead of generating $K$ data points for each of the $U$ iterations, we reuse previous data points and incrementally add a few data points per iteration. 

\section{Selective Data Acquisition}
\label{sec:algorithms}

We first tackle the selective data acquisition problem to minimize model loss and unfairness when slices are independent of each other and then extend our methods to the case where slices may influence each other.

\subsection{Independent Slices}
\label{sec:independentslices}

We first assume that slices do not influence each other. Hence, we only need to solve the optimization problem once. The optimization should be done on all slices as our objective for minimizing loss and unfairness is global. We can formulate a convex optimization problem for selective data acquisition as follows. For the slices $\{s_i\}_{i=1}^n$ with sizes $\{|s_i|\}_{i=1}^n$, we want to find the number of examples to acquire for the slices $\{d_i\}_{i=1}^n$ to minimize the objective function using a total budget of $B$. We assume that the model's loss on a slice follows a power-law curve of the form $b_i (|s_i| + d_i)^{-a_i}$ where $b_i$ and $a_i$ are positive values. We define $
A$ to be \kh{the average loss across all slices} and $C(s_i)$ to be the cost function that captures the effort to acquire an example for $s_i$. The optimization problem is then

%loss of the entire data $D$ 
\[\min \sum_{i=1}^n b_i(|s_i| + d_i)^{-a_i} + \lambda \sum_{i=1}^n \max \left\{0, \frac{b_i(|s_i| + d_i)^{-a_i}}{A} - 1 \right\}\]
\vspace{-0.3cm}
\[
\text{subject to} \sum_{i=1}^n C(s_i) \times d_i = B
\]
where the first term minimizes the loss, and the second term minimizes unfairness by giving a penalty to each slice $s_i$ that has a higher loss than $A$. If the loss of $s_i$ is lower than $A$, we return 0 to prevent the second term from giving a negative value. Since minimizing loss and unfairness are not always aligned, we introduce the $\lambda$ term to balance between minimizing each of them. \kh{By acquiring data for slices with higher losses, we eventually satisfy Definition~\ref{def:unfairness}.}

%\kh{Note that we do not use an absolute difference term (Definition~\ref{def:unfairness}) in the objective function because it does not distinguish low-loss and high-loss slices. Instead, we only want to acquire data for high-loss slices. The acquisition will eventually satisfy Definition ~\ref{def:unfairness} because the absolute difference keeps on decreasing.}

This problem is convex assuming that the model's loss decreases monotonically against more data. The first loss term $b_i(|s_i| + d_i)^{-a_i}$ is a convex function of $d_i$ because it is a power-law curve. The second unfairness term is also convex because $b_i(|s_i| + d_i)^{-a_i}$ is convex, $A$ is a constant for the current $D$, and taking a maximum between two convex functions is convex. Finally, $C(s_i)$ is a constant that varies by slice. Hence, we can derive an optimal solution efficiently using any off-the-shelf convex optimization solver.

% This problem is convex assuming that the model's loss decreases monotonically as more data is acquired. The first loss term $b_i(|s_i| + d_i)^{-a_i}$ is a convex function of $d_i$ because it is a power-law curve. The second unfairness term is also convex because $b_i(|s_i| + d_i)^{-a_i}$ is convex, $A_i$ does not change against $d_i$, and taking a maximum between two convex functions is convex. Finally, $C(s_i)$ is simply a constant that varies by slice. Hence, we can derive an optimal solution efficiently using any off-the-shelf convex optimization solver.

\paragraph*{Algorithm}

The \oneshot{} algorithm updates the learning curves using the techniques in Section~\ref{sec:efficientimplementation} and solves the above optimization problem to determine how much data to acquire for each slice. Note that \oneshot{} always uses the entire budget $B$, assuming the learning curves are perfect, and the slices are independent.
%\kh{and there is no dependency between slices.}

\subsection{Dependent Slices}
\label{sec:dependentslices}

We now consider the case where slices can influence each other. Suppose that there are two slices $s_1$ and $s_2$. If we acquire data for $s_1$ such that it dominates $s_2$ in quantity, then the model training may overfit on $s_1$. Consequently, the model accuracy on $s_2$ may change. Hence, we need to iteratively update the learning curves as more data is acquired. Notice that the iterative updates also serve the dual purpose of making the learning curves more reliable. 

\paragraph*{Modeling the Influence}

%\kh{[[add synthetic exp to show three situations?? ]]} 

\begin{figure}[t]
\begin{subfigure}[t]{0.32\columnwidth}
\centering
\includegraphics[scale=0.6]{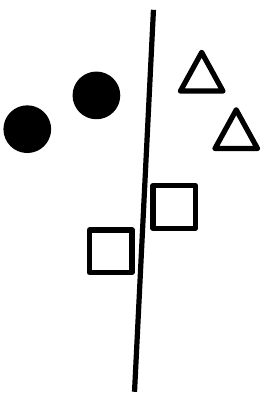}
\caption{}
\label{fig:influences1}
\end{subfigure}
\begin{subfigure}[t]{0.32\columnwidth}
\includegraphics[scale=0.6]{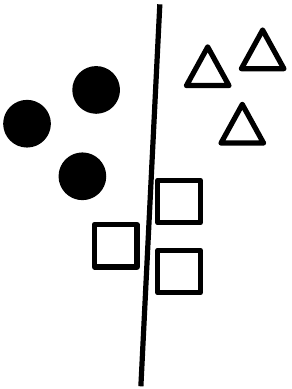}
\caption{}
\label{fig:influences2}
\end{subfigure}
\begin{subfigure}[t]{0.32\columnwidth}
\includegraphics[scale=0.6]{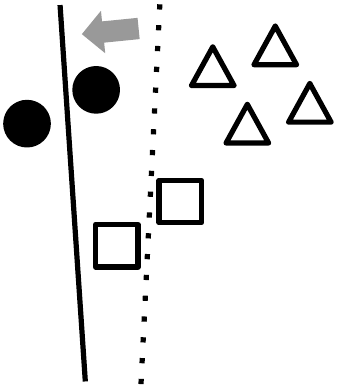}
\caption{}
\label{fig:influences3}
\end{subfigure}
  \vspace{-0.3cm}
\caption{(a) Consider three slices where shape indicates slice, and color indicates label. (b) If we evenly increase the data for all slices, the bias does not change much, and there is little influence among slices. (c) If we only increase the white triangles, the decision boundary may shift to the left due to the new bias, decreasing the loss for the white squares and increasing the loss for the black circles.}
  \label{fig:influences}
  \vspace{-0.4cm}
\end{figure}

We hypothesize that the {\em relative sizes} of the slices (i.e., the bias) and the {\em data similarities} between slices play major roles. Figure~\ref{fig:influences} shows a synthetic example to demonstrate this point. Suppose that we start with three slices (circle, triangle, and square) in Figure~\ref{fig:influences1} where the colors (black and white) indicate the labels. The straight line indicates the decision boundary of the model. Now suppose we evenly increase the data for all slices by 50\% as shown in Figure~\ref{fig:influences2}. Since the bias does not change, the decision boundary does not change much either, which means that there is little influence on any of the slices. However, if we only increase the size of the triangle slice as shown in Figure~\ref{fig:influences3}, suppose the new bias causes the decision boundary to shift towards the left and influences the losses of the square and circle slices. The direction of influence depends on the similarity of data among slices. In our case, the square slice has the same label values as the triangle slice, and its loss decreases as all examples are categorized correctly. On the other hand, the circle slice has different label values than the triangle slice, and its loss increases.

\begin{figure}[t]
\centering
\begin{subfigure}{\columnwidth}
    \includegraphics[width=0.9\columnwidth]{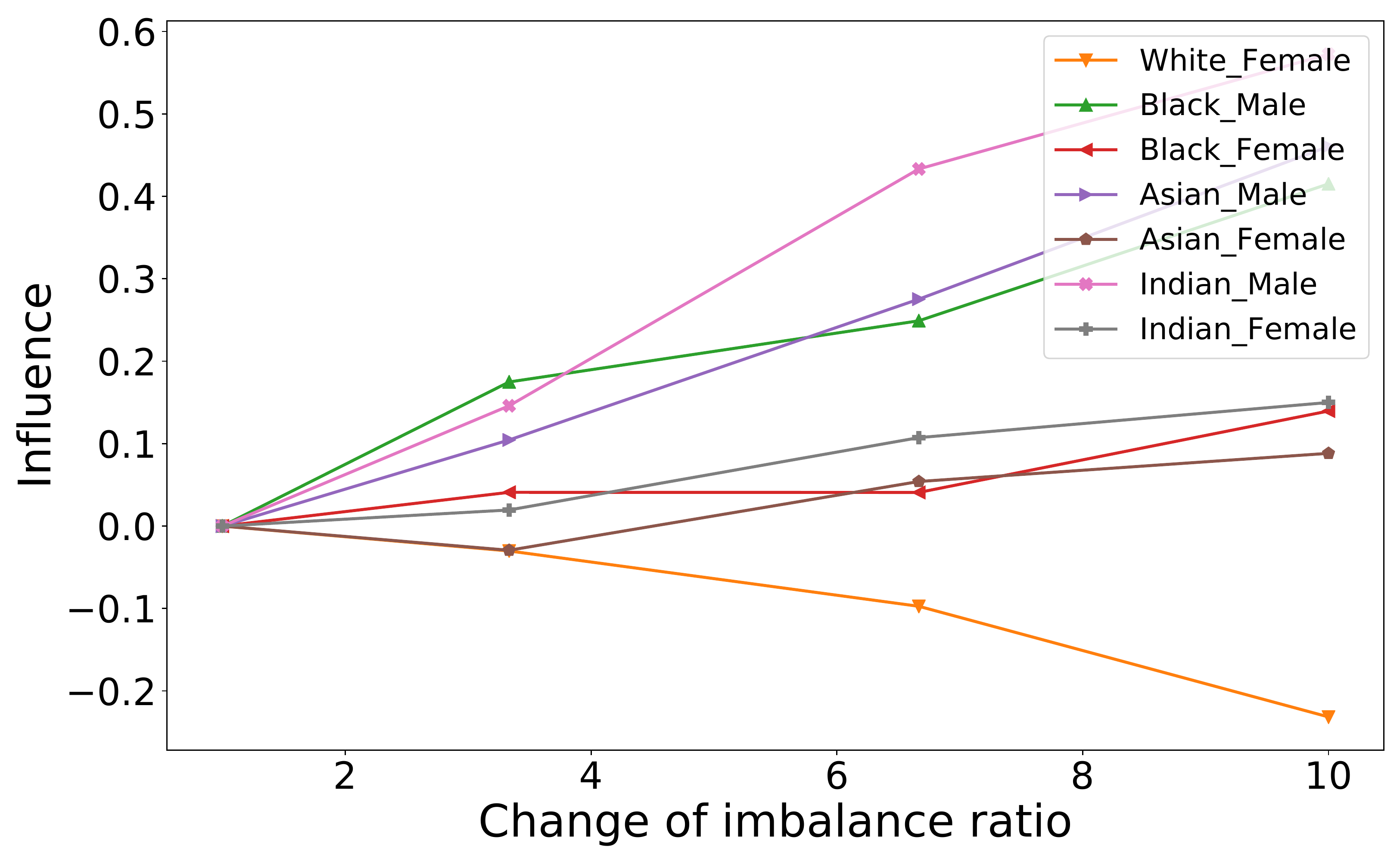}
\end{subfigure}
\vspace{-0.3cm}
\caption{The influences (loss changes) of other UTKFace data slices as more data is acquired for White\_Male, starting from size 300. As the change of the imbalance ratio increases, so does the magnitude of influences on other slices.}
\label{fig:imbalanceratio}
% \vspace{-0.2cm}
\vspace{-0.5cm}
\end{figure}

To verify our hypothesis, we perform an experiment on the real UTKFace dataset~\cite{zhifei2017cvpr} where we use slices that represent different race and gender combinations of people (more details in Section~\ref{sec:experimentalsetting}). We use {\em imbalance ratio}~\cite{BUDA2018249}, which is the maximum ratio between any two slices in $S$, as a proxy for bias. For example, if the slices $s_1$, $s_2$, and $s_3$ have sizes 10, 20, and 30, respectively, then the imbalance ratio is $\frac{max\{10, 20, 30\}}{min\{10, 20, 30\}}$ = $\frac{30}{10}$ = 3. We also define {\em influence} on a slice as the change in loss. According to our hypothesis, if the imbalance ratio changes, the magnitude of influence increases as well. 
We then observe the affect of imbalance ratio on influence as shown in Figure~\ref{fig:imbalanceratio}. Initially, all the slices are of the same size 300, except for the slice White\_Male, which starts from size 50. As we add more examples only to White\_Male, the imbalance ratio change increases, causing the influences on other slices to increase in magnitude as well. While most slices have increasing losses, only the slice White\_Female shows decreasing losses because it has similar data. We make similar observations in other datasets as well.

% \begin{figure}[t]
% \centering
% \begin{subfigure}{\columnwidth}
%     \includegraphics[width=0.95\columnwidth]{slicelossothers_utk_300.pdf}
% \end{subfigure}
% \caption{The losses of 8 UTKFace data slices where data is acquired for the White\_Male slice only. The losses of the other slices are influenced as more examples are acquired.}
% \label{fig:sliceinfluence}
% \end{figure}

%Figure~\ref{fig:sliceinfluence}, but explicitly shows how a positive change in imbalance ratio results in more positive or negative influence. The results for a negative change in imbalance ratio are similar. Hence, we would like to control the imbalance ratio by limiting the amount of data acquired.

\paragraph*{Algorithm}

% \kh{The key idea for handling slice dependencies is to update the learning curves using the point at which the  magnitude of influence increases. Regardless of the influence direction, if the magnitude of influence increases, the learning curves changes as well.} 

\kh{We iteratively update the learning curves whenever "enough" influence occurs, regardless of its direction.} The {\em Iterative} algorithm (shown in Algorithm~\ref{alg:iterative}) limits the change of imbalance ratio to determine how much data to acquire for each slice. We obtain the slice sizes and initialize the imbalance ratio absolute change limit $T$ to 1. While there is enough budget for data acquisition, we increase $T$ per iteration using one of the strategies discussed later. The parameter $L$ specifies the minimum slice sizes to start with and is positive. If any slice is smaller than $L$, we acquire enough examples assuming there is enough budget. (In Section~\ref{sec:unreliablelearningcurves}, we show that $L$ can be a small value.) We then run the \oneshot{} algorithm to derive how many examples to acquire for each slice if we use the entire budget $B$. If the imbalance ratio change would exceed $T$ if we use the entire budget, we limit the number of examples acquired by multiplying $num\_examples$ with the maximum ratio that would not allow that to happen ($change\_ratio$). This problem has nonlinear constraints, and the \textsc{GetChangeRatio} function uses an off-the-shelf optimization library in SciPy to derive a solution. After reflecting $B$, $T$, and $IR$, we repeat the same steps until we run out of budget.

For example, suppose $L = 10$, and there are two slices $s_1$ and $s_2$ with initial sizes of 5 and 10 (i.e., $sizes = [5, 10]$) with a budget of $B = 55$. First, we need to acquire 5 examples for $s_1$ (i.e., $num\_examples$ = [5, 0]) to satisfy $L$ (Step 4). Then we update $sizes$ to $[10, 10]$ and $B$ to 50 (Steps 5--6). Then we set $IR$ to $\frac{10}{10} = 1$. After running \oneshot{}, suppose that $num\_examples$ = [10, 40] (Step 9). If we acquire all this data, the imbalance ratio would become $\frac{10 + 40}{10 + 10} = 2.5$, so $|After\_IR - IR|$ = 2.5 - 1 = 1.5. To avoid exceeding $T = 1$, we compute the change ratio $x$ such that $\frac{10 + 40x}{10 + 10x} = 2$ by invoking the \textsc{GetChangeRatio} function (Step 13). The solution is $x=0.5$, and $num\_examples$ thus becomes $0.5 \times [10, 40] = [5, 20]$. After acquiring the data, we update $sizes$, $B$, $T$, and $IR$ and go back to Step 8.

\begin{algorithm}[t]
    \SetKwInput{Input}{Input}
    \SetKwInOut{Output}{Output}
    \SetNoFillComment
    \SetKwFunction{FIR}{\textnormal{\textsc{GetImbalanceRatio}}}
    \SetKwFunction{GIR}{\textnormal{\textsc{GetIncreaseRatio}}}
    \SetKwProg{Fn}{Function}{:}{}
    \Input{The slices $S$, budget $B$, minimum slice size $L$, and data acquisition cost function $C$}
    %imbalanceratio -> number of training example
    $sizes$ = \textsc{SliceSizes}($S$)\;
    $T$ = 1\;
    \If{$\exists i \  sizes[i] < L$}{
        \tcc{Ensure minimum slice size L}
        $num\_examples$ = $\max(L \times {\bf 1} - sizes, 0)$\;
        $sizes$ = $sizes$ + $num\_examples$\;
        $B$ = $B$ - $\sum_i(C(i) \times num\_examples[i])$\;
        %$T$ = \textsc{GetImbalanceRatio}($sizes$ + $num\_examples$) \;
    }
    $IR$ = \textsc{GetImbalanceRatio}($sizes$)\;
    \While{$B > 0$}{
        \tcc{\oneshot{} always uses the entire budget}
        $num\_examples$ = \textsc{OneShot}($sizes$, $B$)\;
        $After\_IR$ = \textsc{GetImbalanceRatio}($sizes + num\_examples$)\;
        
        \If{$|After\_IR - IR|> T$}{
        \tcc{Do not make imbalance ratio change exceed T}
        $target\_ratio$ = $IR + T \times $ \textsc{Sign}($After\_IR - IR$)\;
        $change\_ratio$ = \textsc{GetChangeRatio}($sizes$, $num\_examples$, $target\_ratio$)\;
        $num\_examples$ = $change\_ratio \times num\_examples$\;
        $After\_IR$ = \textsc{GetImbalanceRatio}($sizes + num\_examples$)\;
        }
        \textsc{CollectData}($num\_examples$)\;
        $sizes$ = $sizes$ + $num\_examples$\;
        $B$ = $B$ - $\sum_i(C(i) \times num\_examples[i])$\;
        $T$ = \textsc{IncreaseLimit}($T$)\;
        $IR$ = $After\_IR$\;
    }
    {\bf return}\;
    %{\bf return} $sizes$ - \textsc{SliceSizes}($S$)\;
    \Fn{\FIR{$sizes$}}{
    {\bf return} $\frac{\max(sizes)}{\min(sizes)}$\;
    }
    \caption{Iterative algorithm for \systems{} }
    \label{alg:iterative}
\end{algorithm}

We now discuss strategies for updating the limit $T$ per iteration using the \textsc{IncreaseLimit} function. On one hand, it is desirable to minimize the number of iterations of Algorithm~\ref{alg:iterative} because each iteration invokes the \oneshot{} algorithm, which involves updating the learning curves and solving an optimization problem. On the other hand, we would like to update the learning curves to be as accurate as possible. While there are many ways to update $T$, we propose the following representative strategies:
\squishlist
    \item \linear{}: For each iteration, leave $T$ as a constant, which limits the imbalance ratio to change linearly. The advantage is that we can avoid mistakenly acquiring too much data due to inaccuracies in the learning curves. However, the number of iterations may be high.
    \item \moderate{}: For each iteration, increase $T$ by a constant $c$. Compared to \linear{}, this approach reduces the number of iterations, but may acquire data unnecessarily.
    \item \exponential{}: For each iteration, multiply $T$ by a constant $c > 1$. Compared to \moderate{}, this strategy collects data even more aggressively using possibly fewer iterations.

    % this approach reduces the number of iterations, but may acquire data unnecessarily.
\squishend

\section{Experiments}
\label{sec:experiments}

In this section, we evaluate \systems{} on real datasets and address the following questions:

\squishlist
    \item How reliable and efficient is the learning curve generation used in \systems{}?
    \item How does \systems{} compare with the baselines in terms of model loss and unfairness?
    \item How does \systems{} perform on small slices where the learning curves are unreliable?
\squishend

\systems{} is implemented in TensorFlow~\cite{DBLP:conf/osdi/AbadiBCCDDDGIIK16} and Keras, and we use Titan RTX GPUs for model training.

\subsection{Setting}
\label{sec:experimentalsetting}

\paragraph*{Datasets} We experiment on the following four datasets that capture different characteristics of the slices. 
%Figure~\ref{fig:datasets} shows samples of the image datasets. 
While the AdultCensus dataset is the most widely-used in the fairness literature, \systems{} is not limited to any particular dataset because our unfairness measure (Definition~\ref{def:unfairness}) does not require sensitive attributes (e.g., race and gender).

\squishlist
    % \item UCI Adult Dataset~\cite{DBLP:conf/kdd/Kohavi96}: Contains people records containing features including age, education, occupation, native country, and sex. The prediction task is to determine if a person makes over \$50K per year. The slicing was done on the native country attribute. Depending on the country, the learning curve may differ significantly.
    % \item MNIST~\cite{lecun-01a}: Contains images that represent digits from 0 to 9 where the goal is to predict the digit of each image. Here we slice the images according to their labels, i.e., there are 10 slices in total. Compared to the other datasets, the slices are the most ``homogeneous'' in the sense that they are all about digits.
    % \item Fashion-MNIST~\cite{xiao2017/online}: Contains images that can be categorized as one of 10 type of clothes, e.g., shoes, shirts, pants, and more. Compared to MNIST, the slices have more variety in terms of the objects they represent.
    
    \item Fashion-MNIST~\cite{xiao2017/online}: Contains images that can be categorized as one of 10 type of clothes, e.g., shoes, shirts, pants, and more. Here we slice the images according to their labels, i.e., there are 10 slices in total.
    \item Mixed-MNIST: Combines the Fashion-MNIST dataset and MNIST dataset\cite{lecun-01a}, which contains images that represent digits from 0 to 9. Created to demonstrate a dataset with a large number (20) of non-homogeneous slices from two data sources.
    \item UTKFace~\cite{zhifei2017cvpr}: Contains various face images of people of different (male and female) gender and race (White, Black, Asian, and Indian), used for race classification. We used 8 slices by combining two genders and four races, e.g., Black female. 
    %It is known that for some of these slices, the classification loss is higher than other slices~\cite{das:hal-01892103}. 
    \item AdultCensus~\cite{DBLP:conf/kdd/Kohavi96}: Contains people records with features including age, education, and sex. Used to predict which people make over \$50K per year. We use 4 slices by combining two races (White and Black) and two genders (male and female).
\squishend

% \begin{figure}[t]
% \centering
% \begin{subfigure}{\columnwidth}
%      \includegraphics[width=\columnwidth]{dataset-fashion.pdf}
%      \caption{Fashion-MNIST dataset}
% \end{subfigure} 
% \begin{subfigure}{\columnwidth}
%      \includegraphics[width=\columnwidth]{dataset-mixed.pdf}
%      \caption{Mixed-MNIST dataset}
% \end{subfigure}
% \begin{subfigure}{\columnwidth}
%      \includegraphics[width=\columnwidth]{dataset-utkface.pdf}
%      \caption{UTKFace dataset}
% \end{subfigure} 
% \caption{Samples of the three image datasets.}
% \label{fig:datasets}
% \vspace{-0.2cm}
% \end{figure}

\paragraph*{Data Acquisition and Cost Function} For the Fashion-MNIST, Mixed-MNIST, and AdultCensus datasets, we first simulate data acquisition by starting from a subset and adding more examples. This approach is reasonable because we are not tied to any data acquisition technique. We also define the cost function to always return 1. For the UTKFace dataset, we use a real scenario where we crowdsource new images using Amazon Mechanical Turk (AMT)~\cite{amazonmechanicalturk} and store them in Amazon S3. We design a task by asking a worker to find new face images of a certain demographic (e.g., $slice = White\_Women$) from any website. We pay 4 cents per image found, employ workers from all possible countries, and acquired images during 8 separate time periods. We do not show the workers all the images acquired so far, so they may acquire duplicate images. However, the duplicate rate is not as high as one may think because workers around the world use a wide range of websites to acquire images. Some workers make mistakes and acquire incorrect images that do not fit in the specified demographic. Hence, we include a post-processing step of filtering obvious errors manually, removing exact duplicates, and cropping faces using Google Cloud AI Platform services. We also define the collection cost of a slice to be proportional to the average time a task is finished. Table~\ref{tbl:costfunction} shows the average time (seconds) to acquire images for the 8 UTKFace slices. Interestingly, the collection costs can be quite different. For example, an Indian woman image takes 50\% longer to acquire than a Black male image and thus has a cost of 1.5.

\begin{table}[ht]
  \centering
  \begin{tabular}{@{\hspace{1pt}}c@{\hspace{3pt}}c@{\hspace{3pt}}c@{\hspace{3pt}}c@{\hspace{3pt}}c@{\hspace{3pt}}c@{\hspace{3pt}}c@{\hspace{3pt}}c@{\hspace{3pt}}c@{\hspace{1pt}}}
    \toprule
   & $W\_M$ & $W\_F$ & $B\_M$ & $B\_F$ & $A\_M$ & $A\_F$ & $I\_M$ & $I\_F$ \\
    \midrule
   Avg. time (s) &  82.1 & 81.9 & 67.6 & 79.3 & 94.8 & 77.5 & 91.6 & 104.6 \\
   Cost $C$ & 1.2 & 1.2 & 1 & 1.2 & 1.4 & 1.1 & 1.4 & 1.5 \\
    \bottomrule
  \end{tabular}
  \caption{The collection costs of UTKFace slices are proportional to the average times to finish tasks. Here $W$ = White, $B$ = Black, $I$ = Indian, $A$ = Asian, $M$ = Male, and $F$ = Female.}
  \label{tbl:costfunction}
  \vspace{-0.8cm}
%   \vspace{-1cm}
\end{table}

\paragraph*{Methods Compared} We compare the following methods:

\squishlist
    \item {\em Uniform (baseline 1)}: collects similar amounts of data per slice.
    \item {\em Water filling (baseline 2)}: collects data such that the slices end up having similar amounts of data.
    \item \oneshot{}: updates the learning curves and solves the optimization problem once and collects data as described in Section~\ref{sec:independentslices}. As a default, we set $\lambda = 1$.
    \item {\em Iterative}: iteratively updates the learning curves and collects data as described in Section~\ref{sec:dependentslices}. We use three iteration strategies for increasing $T$ per iteration: \linear{} (fixes $T$ to 1), \moderate{} (increases $T$ by 1), and \exponential{} (multiplies $T$ by 2). 
    %The \linear{} strategy shows identical results to \exponential{} in all our experiments.
\squishend

\kh{We use the efficient implementation method (Section~\ref{sec:efficientimplementation}) to fit learning curves for all experiments.}

\paragraph*{Measures} We use the model loss and unfairness measures based on our discussions in Section~\ref{sec:preliminaries}. For all measures, lower values are better. We report mean values over 10 trials for all measures.
\squishlist
    \item {\em Loss}: the average log loss for multi-class classification.
    \item {\em Average Equalized Error Rates (Avg. EER)}: the unfairness measure in Definition~\ref{def:unfairness}, i.e., the avg.\@ loss difference between a slice and the entire data.
    \item {\em Maximum Equalized Error Rates (Max. EER)}: same as avg.\@ EER, except that we take the maximum loss difference instead of the average. Used to understand the worst-case unfairness.
\squishend

For each slice, we split the available data into train and validation sets and measure the loss on the validation set. We set the validation set size to be 500 per slice.

\paragraph*{Models and Hyperparameter Tuning}

For the Fashion-MNIST, Mixed-MNIST, and UTKFace datasets, we use basic convolutional neural networks with 2--3 hidden layers. For the AdultCensus dataset, we train a fully connected network with no hidden layers. For all datasets, we initially set the hyperparameters using simple grid search. Afterwards, we do not further change the hyperparameters while running \systems{} for consistent model training.

\subsection{Learning Curve Analysis}

\begin{figure*}[htbp]
\centering
%      \includegraphics[scale=0.65]{LearningcurveExamples.pdf}
% \caption{The learning curves of the MNIST, Fashion-MNIST, and UTKFace datasets. For each dataset, we show two learning curves for different slices on the same column. Even for seemingly-even slices in the MNIST dataset, the learning curves can be quite different.}
% \begin{subfigure}{0.245\textwidth}
%      \includegraphics[width=\textwidth]{lc_mnist_two.pdf}
%      \caption{MNIST}
% \end{subfigure}
\begin{subfigure}{0.245\textwidth}
     \includegraphics[width=\textwidth]{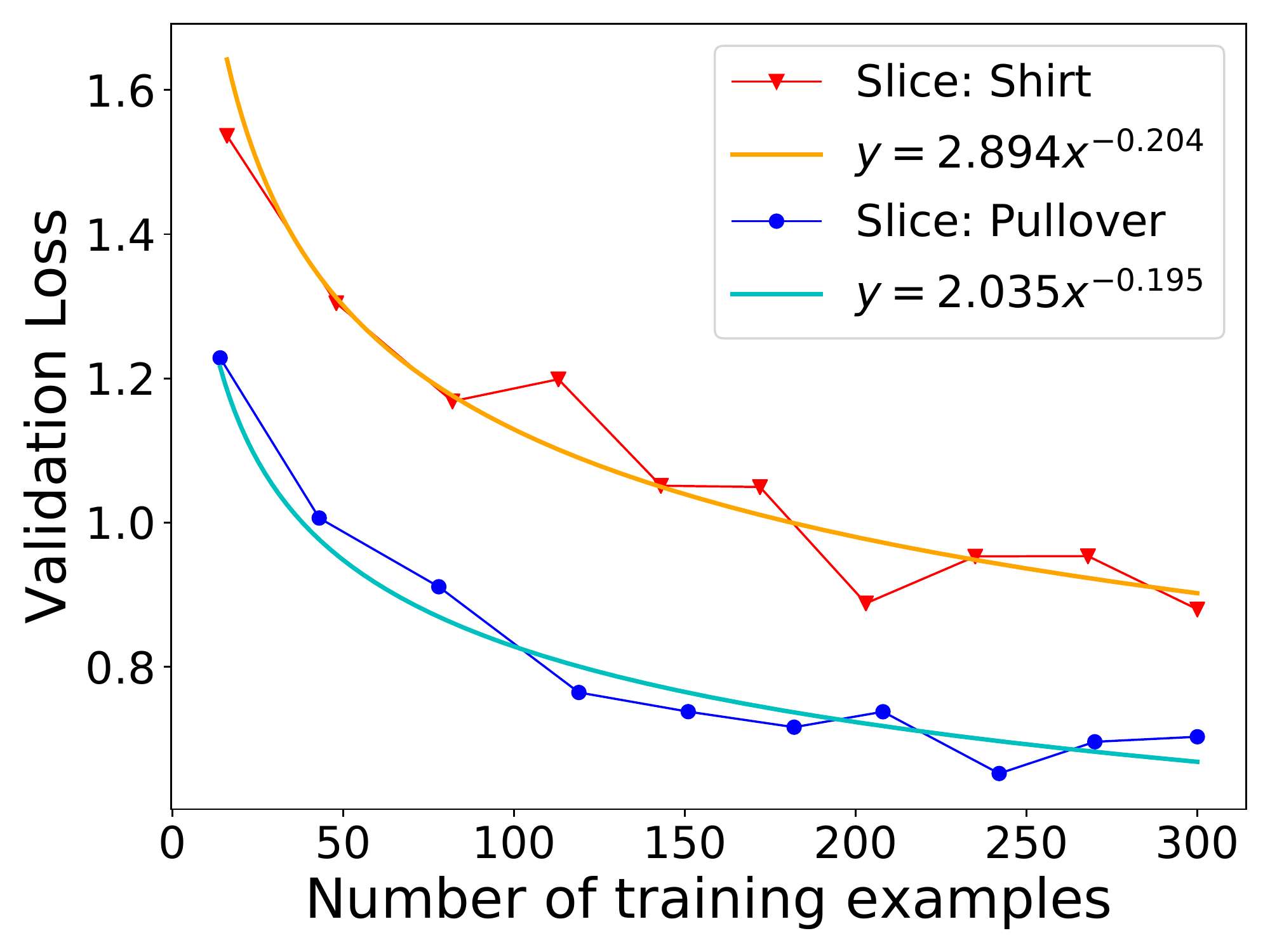}
     \caption{Fashion-MNIST}
\end{subfigure} 
\begin{subfigure}{0.245\textwidth}
     \includegraphics[width=\textwidth]{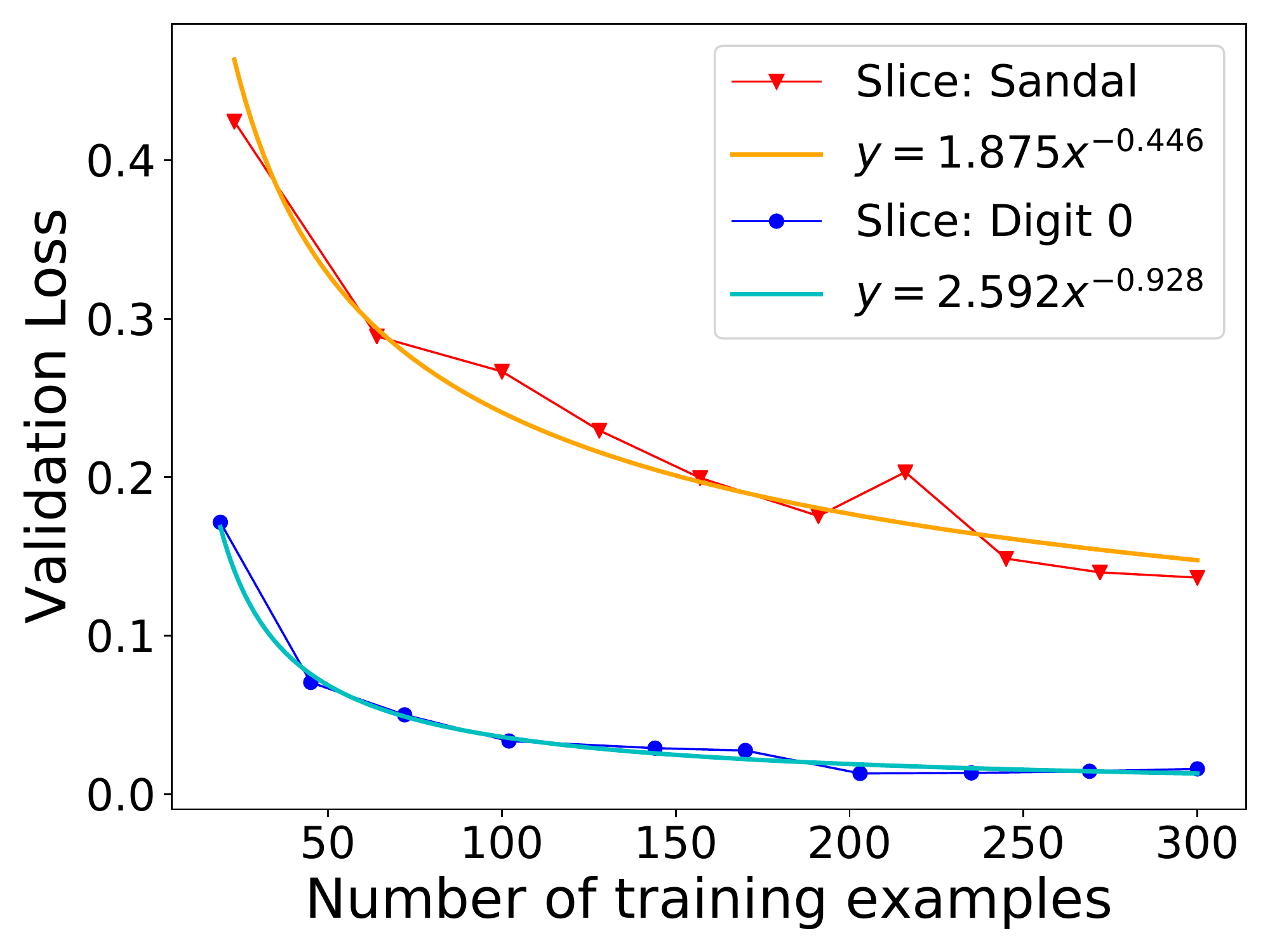}
     \caption{Mixed-MNIST}
\end{subfigure} 
\begin{subfigure}{0.245\textwidth}
     \includegraphics[width=\textwidth]{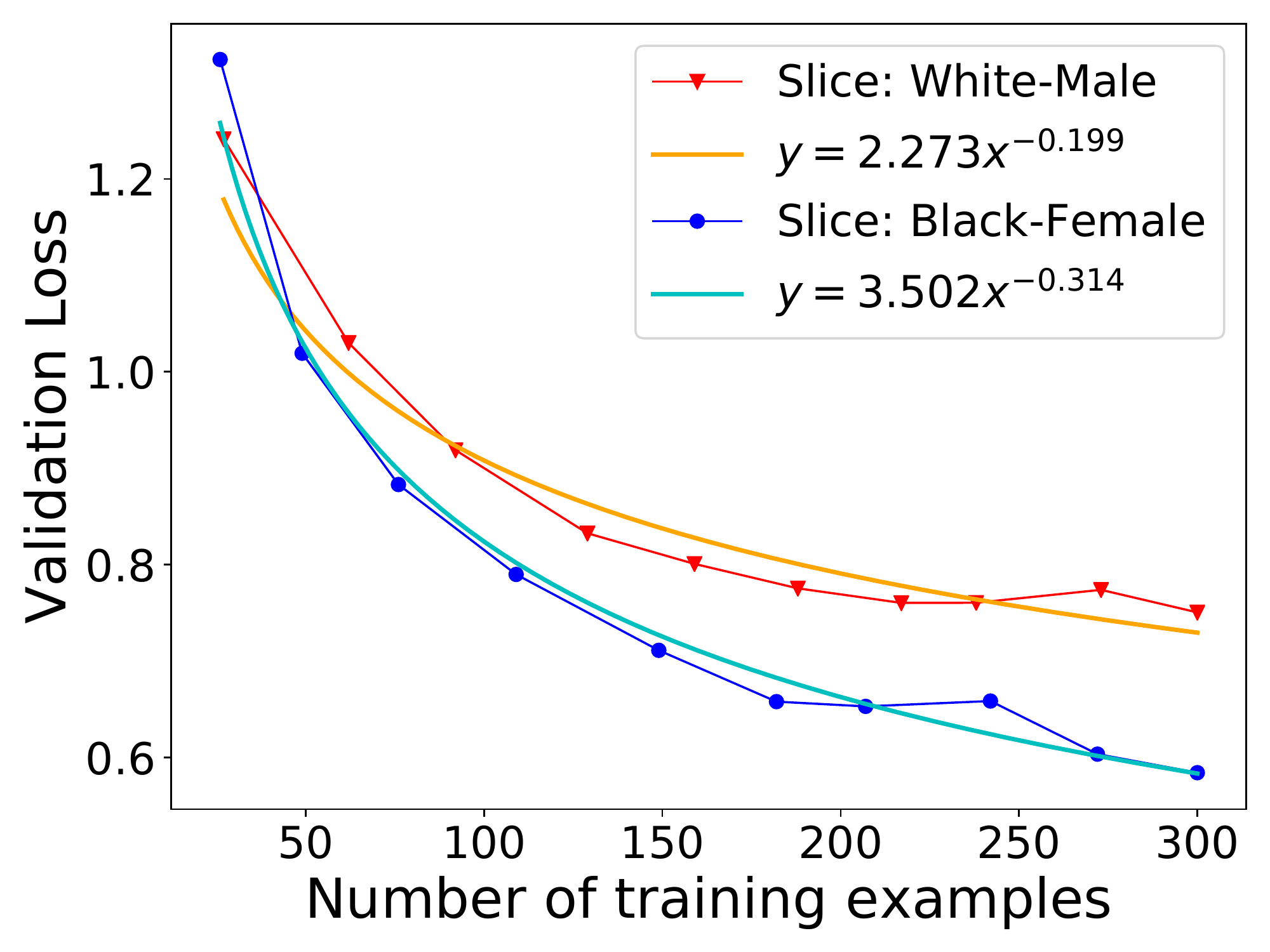}
     \caption{UTKFace}
\end{subfigure} 
\begin{subfigure}{0.245\textwidth}
     \includegraphics[width=\textwidth]{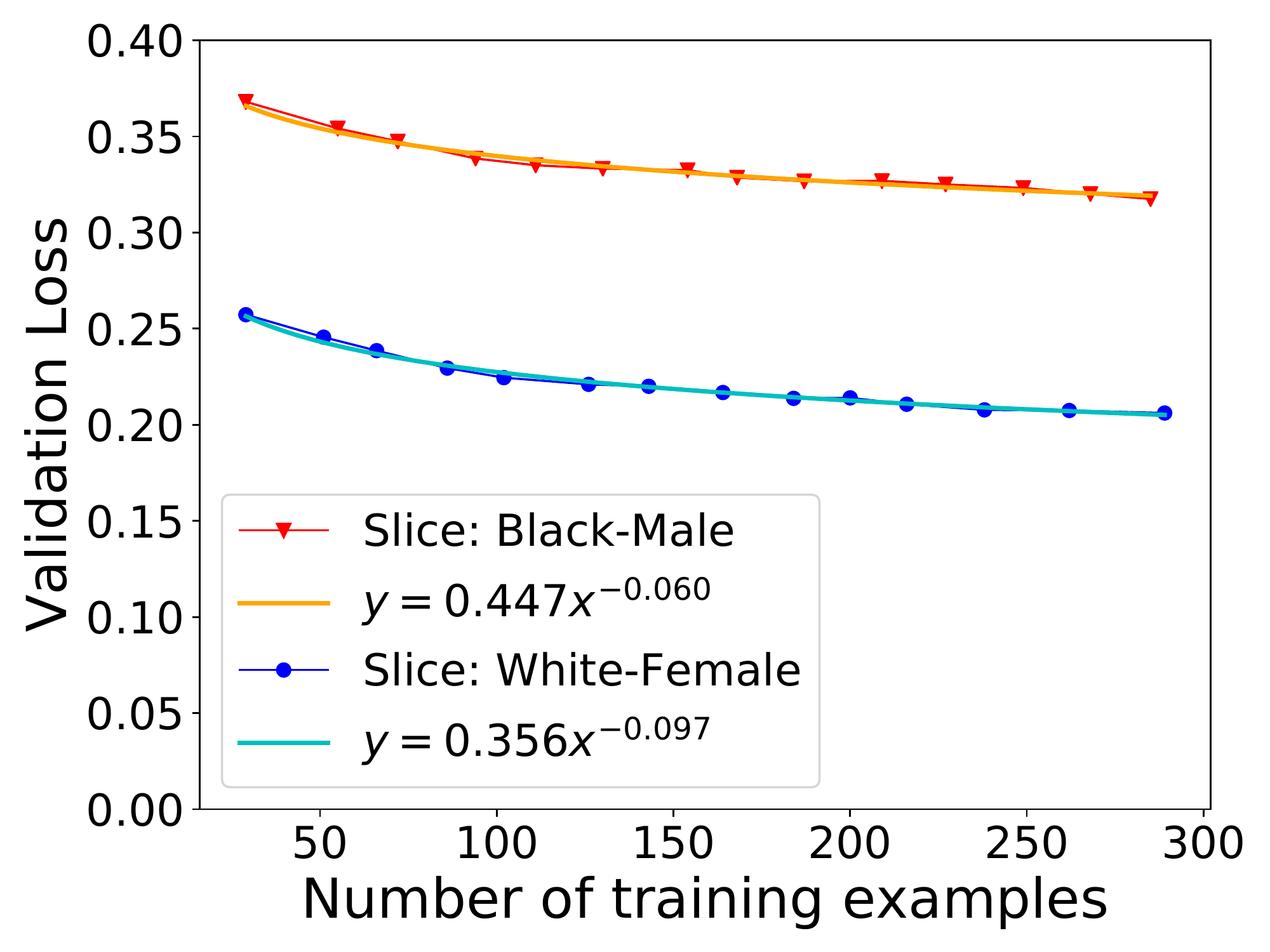}
     \caption{AdultCensus}
\end{subfigure} 
\vspace{-0.2cm}
\caption{The learning curves of the four datasets. The initial slice sizes are set to be large enough so we only observe power-law regions without small-data regions. For each dataset, we show two learning curves for different slices fitted using a validation set. Even for seemingly-homogeneous slices in the Fashion-MNIST dataset, the learning curves can be quite different.} %\sw{combine the two graphs for each dataset into one}}
\vspace{-0.2cm}
\label{fig:mnistcurves}
\end{figure*}

We analyze our learning curve estimation method described in Section~\ref{sec:learningcurve}. For each slice, we take $K = 10$ random samples of the data with differing sizes and use a non-linear least squares method to fit a curve. Figure~\ref{fig:mnistcurves} shows two learning curves for each dataset where the x-axis is the subset size, and the y-axis is the loss on a validation set. Here we only observe power-law regions and not the small-data regions (see Figure~\ref{fig:curveanalysis}) because the initial slice sizes are large enough. As the subset size increases, the loss decreases as well. Even for the most homogeneous dataset Fashion-MNIST, the learning curves can be different, resulting in different amounts of data acquisition for the slices.  

%If we decrease the initial slice sizes, then we can see the small data region in Figure~\ref{fig:curveanalysis}.

\begin{figure}[t]
\centering
  \includegraphics[width=0.8\columnwidth]{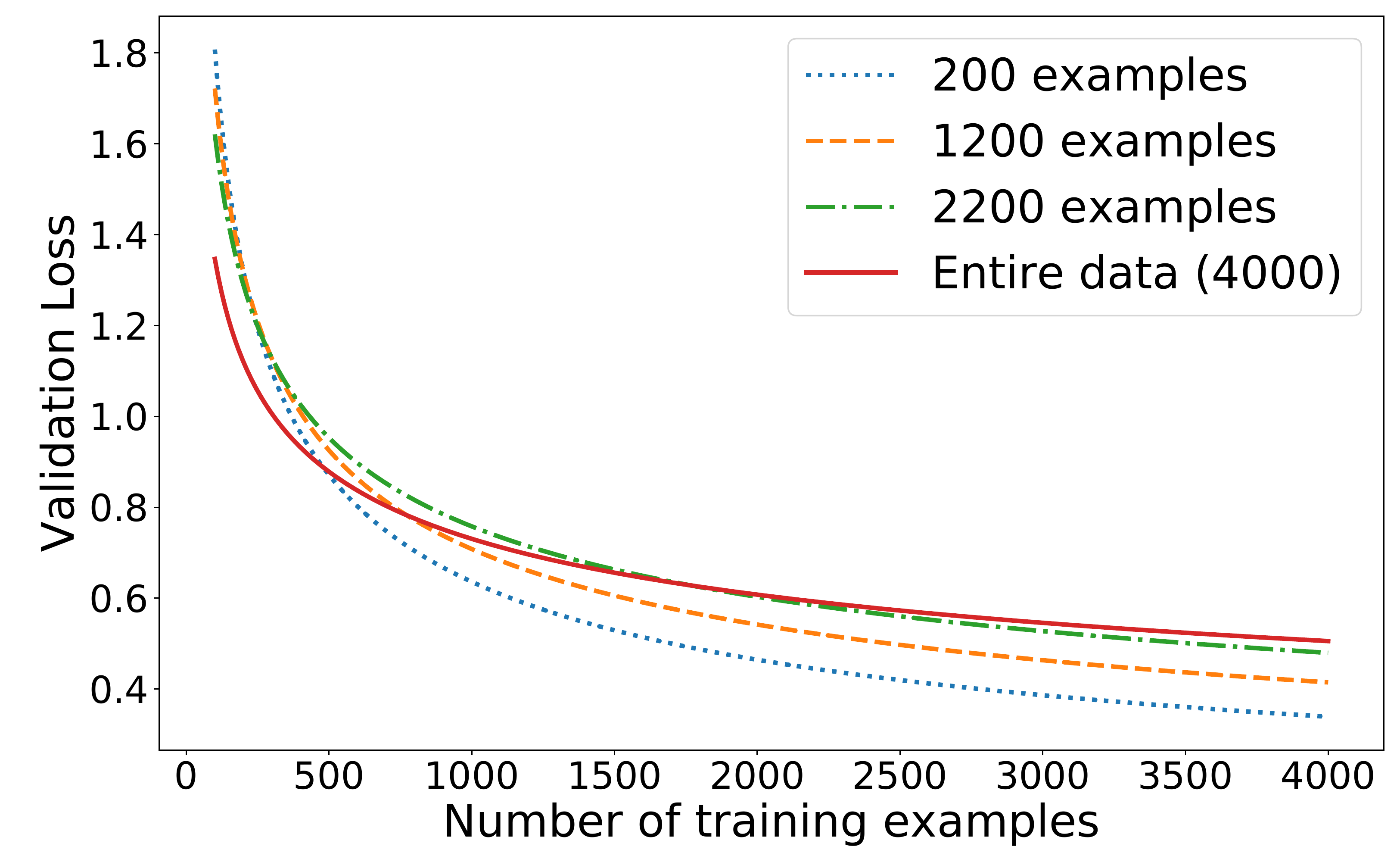}
  \vspace{-0.2cm}
  \caption{Given a slice in the Fashion-MNIST dataset, we increase its size and train new learning curves. As a result, the learning curves trained on small slices deviate more from the other curves and can be viewed as erroneous.}
  \label{fig:mnistcurvesaccuracy}
%   \vspace{-0.2cm}
  \vspace{-0.2cm}
\end{figure}

We also observe how the learning curve changes as the slice itself grows in size. That is, we increase the slice size and, for each size, fit a new learning curve. Figure~\ref{fig:mnistcurvesaccuracy} shows the learning curves for a slice in the Fashion-MNIST dataset. As a result, the smaller the slice, the more the learning curve deviates from the others. This result shows that the learning curves must be updated as more data is acquired, especially for slices that start small.

\subsection{Selective Data Acquisition Optimization}
\label{sec:selectivedatacollectionexperiments}

We now show the loss and unfairness results of \systems{} and make a comparison with the baselines.

\subsubsection{Loss and Unfairness of \systems{} Methods}
\label{sec:lossunfairness}

Table~\ref{tbl:systemsmeasures} compares the loss and unfairness results of the \systems{} methods on the four datasets. {\em Original} is where we train a model on the current slices with no data acquisition. As a result, the \systems{} methods improve {\em Original} both in terms of loss and unfairness. Among the \systems{} methods, the iterative methods outperform \oneshot{}. Between the \exponential{} and \linear{} methods, \linear{} usually has lower loss and unfairness than \exponential{}. This result is expected because \linear{} uses more iterations to update the learning curves to be more reliable. Finally, \moderate{} and \exponential{} perform similarly, although \moderate{} has lower loss and unfairness on the Fashion-MNIST dataset.

% \begin{table}[t]
%   \centering
%   \begin{tabular}{cccc}
%     \toprule
%     {\bf Dataset} & {\bf Method} & {\bf Loss} & {\bf Avg./Max. EER} \\ %& {\bf \# Iters} \\
%     \midrule
%     \multirow{4}{*}{MNIST} & Original & 0.241 & 0.071 / 0.100 \\ %& n/a \\
%     & \oneshot{} & 0.178 & 0.037 / 0.057 \\ %& 1 \\
%     & \exponential{} & 0.167 & 0.028 / 0.058 \\ %& 3 \\
%     & \linear{} & {\bf 0.154} & {\bf 0.014} / {\bf 0.031} \\ %& 5 \\
%     \midrule
%     \multirow{4}{*} & Original & 0.626 & 0.436 / 0.905 \\ %& n/a \\
%     {Fashion-}& \oneshot{} & 0.488 & 0.239 / {\bf 0.351} \\ %& 1 \\
%     {MNIST}& \exponential{} & {\bf 0.474} & 0.215 / 0.482 \\ %& 3 \\
%     & \linear{} & {\bf 0.474} & {\bf 0.212} / 0.423 \\ %& 5 \\
%     \midrule
%     \multirow{4}{*}{UTKFace} & Original & 0.886 & 0.121 / 0.217 \\ %& n/a \\
%     & \oneshot{} & 0.691 & 0.087 / {\bf 0.119} \\ %& 1 \\
%     & \exponential{} & {\bf 0.681} & {\bf 0.050} / 0.134 \\ %& 2 \\
%     & \linear{} & {\bf 0.681} & {\bf 0.050} / 0.134 \\ %& 3 \\
%     \midrule
%     \multirow{4}{*}{AdultCensus} & Original & 0.303 & 0.155 / 0.270 \\ %& n/a \\
%     & \oneshot{} & 0.289 & 0.151 / 0.263 \\ %& 1 \\
%     & \exponential{} & {\bf 0.281} & {\bf 0.147} / {\bf 0.252} \\ %& 2 \\
%     & \linear{} & {\bf 0.281} & {\bf 0.147} / {\bf 0.252} \\ %& 3 \\
%     \bottomrule
%   \end{tabular}
%   \caption{\systems{} methods comparison on the 4 datasets.}
%   \label{tbl:systemsmeasures}
%   %\vspace{-0.4cm}
% \end{table}

\begin{table}[t]
  \centering
  \begin{tabular}{cccc}
    \toprule
    {\bf Dataset} & {\bf Method} & {\bf Loss} & {\bf Avg./Max. EER} \\ %& {\bf \# Iters} \\
    % \midrule
    % \multirow{5}{*}{MNIST} & Original & 0.310 & 0.110 / 0.172 \\ %& n/a \\
    % & \oneshot{} & 0.202 & 0.076 / 0.157 \\ %& 1 \\
    % & \exponential{} & 0.201 & 0.076 / 0.149 \\ %& 3 \\
    % & \moderate{} & 0.201 & 0.076 / 0.149 \\ %& 3 \\
    % & \linear{} & {\bf 0.172} & {\bf 0.033} / {\bf 0.075} \\ %& 5 \\
    \midrule
    \multirow{5}{*}{\begin{tabular}{@{}c@{}}Fashion- \\ MNIST\end{tabular}} & Original & 0.423 & 0.255 / 0.617 \\ %& n/a \\
    & \oneshot{} & 0.327 & 0.174 / 0.542 \\ %& 1 \\
    & \exponential{} & 0.304 & 0.144 / 0.351 \\ %& 3 \\
    & \moderate{} & 0.302 & {\bf 0.134} / 0.319 \\ %& 3 \\
    & \linear{} & {\bf 0.300} &{\bf 0.134} / {\bf 0.313} \\ %& 5 \\
    \midrule
    % \multirow{5}{*}{\begin{tabular}{@{}c@{}}Mixed- \\ MNIST\end{tabular}} & Original & 0.288 & 0.196 / 0.670 \\ %& n/a \\
    % & \oneshot{} & 0.229 & 0.118 / {\bf0.355} \\ %& 1 \\
    % & \exponential{} & 0.225 & {\bf0.112} / 0.417 \\ %& 3 \\
    % & \moderate{} & {\bf0.223} & {\bf0.112} / 0.494 \\ %& 3 \\
    % & \linear{} & {\bf0.223} & {\bf0.112} / 0.484 \\ %& 5 \\
    % \midrule
    \multirow{5}{*}{\begin{tabular}{@{}c@{}}Mixed- \\ MNIST\end{tabular}} & Original & 0.299 & 0.198 / 0.691 \\ %& n/a \\
    & \oneshot{} & 0.266 & 0.181 / 1.169 \\ %& 1 \\
    & \exponential{} & 0.225 & 0.119 / \bf{0.395} \\ %& 3 \\
    & \moderate{} & 0.227 & 0.115 / 0.398 \\ %& 3 \\
    & \linear{} & {\bf0.223} & {\bf0.111} / 0.489 \\ %& 5 \\
    \midrule
    \multirow{5}{*}{UTKFace} & Original & 0.590 & 0.097 / 0.207 \\ %& n/a \\
    & \oneshot{} & 0.578 & 0.077 / 0.197 \\ %& 1 \\
    & \exponential{} & \bf{0.572} & \bf{0.069} / \bf{0.184} \\ %& 3 \\
    & \moderate{} & \bf{0.572} & \bf{0.069} / \bf{0.184} \\ %& 3 \\
    & \linear{} & \bf{0.572} & \bf{0.069} / \bf{0.184} \\ %& 3 \\
    \midrule
    \multirow{5}{*}{AdultCensus} & Original & 0.264 & 0.104 / 0.168 \\ %& n/a \\
    & \oneshot{} & 0.253 & 0.094 / 0.150 \\ %& 1 \\
    & \exponential{} & 0.252 & 0.094 / {\bf0.144} \\ %& 3 \\
    & \moderate{} & 0.252 & 0.094 / {\bf0.144} \\ %& 3 \\
    & \linear{} & {\bf 0.251} & {\bf 0.093} / {\bf 0.144} \\ %& 5 \\
    \bottomrule
  \end{tabular}
  \caption{\systems{} methods comparison on the 4 datasets.}
  \label{tbl:systemsmeasures}
  \vspace{-1.1cm}
%   \vspace{-0.75cm}
\end{table}

\begin{table*}[t]
  \centering
  \begin{tabular}{ccccccccccccc}
    \toprule
   {\bf Dataset} & {\bf Method} & {\bf 0} & {\bf 1} & {\bf 2} & {\bf 3} & {\bf 4} & {\bf 5} & {\bf 6} & {\bf 7} & {\bf 8} & {\bf 9} & {\bf \# iterations} \\
    % \midrule
    % \multirow{5}{*}{\begin{tabular}{@{}c@{}}MNIST \\ ($B = 8K$)\end{tabular}} & Original & 300 & 300 & 300 & 300 & 300 & 300 & 300 & 300 & 300 & 300 & n/a \\
    % & \oneshot{} & 112 & 281 & 2305 & 523 & 488 & 611 & 1 & 87 & 2394 & 1198 & 1 \\
    % & \exponential{} & 50 & 102 & 1690 & 637 & 404 & 594 & 0 & 733 & 1802 & 1988 & 3 \\
    % & \moderate{} & 50 & 102 & 1690 & 637 & 404 & 594 & 0 & 733 & 1802 & 1988 & 3 \\
    % & \linear{} & 89 & 422 & 1608 & 978 & 349 & 810 & 0 & 621 & 1646 & 1477 & 6 \\
    \midrule
    \multirow{5}{*}{\begin{tabular}{@{}c@{}}Fashion-MNIST \\ ($B = 6K$)\end{tabular}} & Original & 200 & 200 & 200 & 200 & 200 & 200 & 200 & 200 & 200 & 200 & n/a \\
    & \oneshot{} & 76 & 40 & 761 & 397 & 2203 & 177 & 1829 & 268 & 247 & 1 & 1 \\
    & \exponential{} & 888 & 59 & 1373 & 693 & 1290 & 40 & 1400 & 112 & 145 & 0 & 4\\
    & \moderate{} & 657 & 45 & 1336 & 755 & 1239 & 36 & 1701 & 96 & 135 & 0 & 4 \\
    & \linear{} & 671 & 3 & 1296 & 599 & 1495 & 40 & 1627 & 82 & 173 & 14 & 9 \\
    % \midrule
    % \multirow{5}{*}{\begin{tabular}{@{}c@{}}Mixed-MNIST \\ ($B = 6K$)\end{tabular}} & Original & 150 & 150 & 150 & 150 & 150 & 150 & 150 & 150 & 150 & 150 & n/a \\
    % & \oneshot{} & 0 & 0 & 73 & 0 & 95 & 473 & 1313 & 487 & 1225 & 1953 & 1\\
    % & \exponential{} & 0 & 0 & 49 & 0 & 97 & 1152 & 1045 & 555 & 1430 & 1274 & 4\\
    % & \moderate{} & 0 & 0 & 61 & 0 & 100 & 859 & 1400 & 589 & 1345 & 1267 & 4\\
    % & \linear{} & 0 & 1 & 39 & 0 & 126 & 1035 & 1400 & 634 & 1077 & 1250 & 8 \\
    \midrule
    \multirow{5}{*}{\begin{tabular}{@{}c@{}}Mixed-MNIST \\ ($B = 6K$)\end{tabular}} & Original & 150 & 150 & 150 & 150 & 150 & 150 & 150 & 150 & 150 & 150 & n/a \\
    & \oneshot{} & 33 & 0 & 166 & 47 & 120 & 1092 & 0 & 637 & 521 & 2557 & 1 \\
    & \exponential{} & 1 & 16 & 64 & 7 & 163 & 933 & 818 & 597 & 1202 & 1552 & 4 \\
    & \moderate{} & 1 & 0 & 63 & 5 & 99 & 738 & 1225 & 744 & 989 & 1500 & 5 \\
    & \linear{} & 1 & 2 & 68 & 6 & 133 & 790 & 1076 & 736 & 1200 & 1360 & 10 \\
    \midrule
    \multirow{5}{*}{\begin{tabular}{@{}c@{}}UTKFace \\ ($B = 3K$)\end{tabular}} & Original & 400 & 400 & 400 & 400 & 400 & 400 & 400 & 400 & - & - & n/a \\
    & \oneshot{} & 470 & 427 & 242 & 519 & 448 & 311 & 1263 & 238 & - & - & 1\\
    & \exponential{} & 667 & 384 & 158 & 489 & 560 & 337 & 930 & 365 & - & - & 2\\
    & \moderate{} & 667 & 384 & 158 & 489 & 560 & 337 & 930 & 365 & - & - & 2\\
    & \linear{} & 667 & 384 & 158 & 489 & 560 & 337 & 930 & 365 & - & - & 2\\
    \midrule
    \multirow{5}{*}{\begin{tabular}{@{}c@{}}AdultCensus \\ ($B = 500$)\end{tabular}} & Original & 150 & 150 & 150 & 150 & - & - & - & - & - & - & n/a \\
    & \oneshot{} & 0 & 0 & 55 & 445 & - & - & - & - & - & - & 1\\
    & \exponential{} & 3 & 0 & 164 & 333 & - & - & - & - & - & - & 2\\
    & \moderate{} & 3 & 0 & 164 & 333 & - & - & - & - & - & - & 2\\
    & \linear{} & 1 & 0 & 170 & 329 & - & - & - & - & - & - & 3 \\
    \bottomrule
  \end{tabular}
  \caption{Selective data acquisition results for the experiments in Table~\ref{tbl:systemsmeasures} where slices are listed numbered from 0 up to 9 (the ordering is not important). For Mixed-MNIST, we select 10 out of the 20 slices. The {\em Original} row contains the initial number of examples for each slice while the other rows show the additional number of examples acquired.}
  \label{tbl:systemscollection}
%   \vspace{-0.5cm}
  \vspace{-0.5cm}
\end{table*}

Table~\ref{tbl:systemscollection} shows how much data is acquired for each method in Table~\ref{tbl:systemsmeasures} for the four datasets. For each dataset, the initial slice sizes (same as $L$) are specified in the {\em Original} row. While \oneshot{} has only one chance to decide how much data to collect, the \exponential{}, \moderate{}, and \linear{} methods have more chances to adjust their results. For example, on the Fashion-MNIST dataset, \oneshot{} overshoots and collects too much data for slices \#4 and \#6 while the other methods properly adjust their learning curves through more iterations. Another observation is that \linear{} uses more iterations than \moderate{} and \exponential{} because it is more conservative in increasing $T$. The exception is UTKFace where all methods perform two iterations. Hence, \linear{} can be viewed as trading off efficiency for the lower loss and unfairness results in Table~\ref{tbl:systemsmeasures}. 

% We also perform the above experiments when the initial slice sizes are not the same and follow an exponential distribution in our technical report~\cite{slicetunertr}. As a result, we make similar observations regarding the \systems{} performances.

We also perform the above experiments when the initial slice sizes are not the same and follow an exponential distribution (see Appendix ~\ref{sec:exponential}). As a result, we make similar observations regarding the \systems{} performances.

In summary, the iterative algorithms clearly outperform \oneshot{}. Also, while \moderate{} and \exponential{} have slightly-worse loss and unfairness than \linear{}, they use much fewer iterations. Finally, \moderate{} performs similar to \exponential{} overall, but better on Fashion-MNIST. In the following sections, we thus only use \moderate{} as a representative strategy.

\begin{table}[t]
  \centering
  \begin{tabular}{cccc}
    \toprule
    {\bf Dataset} & {\bf $\lambda$} & {\bf Loss} & {\bf Avg./Max. EER} \\
    \midrule
    \multirow{4}{*}{Fashion-MNIST} & 0 & 0.284 & 0.160 / 0.402 \\
    & 0.1 & 0.285 & 0.148 / 0.330 \\
    & 1 & 0.302 & 0.129 / 0.330 \\
    & 10 & 0.317 & 0.112 / 0.217 \\
    \midrule
    \multirow{4}{*}{Mixed-MNIST} & 0 & 0.208 & 0.136 / 0.582\\
    & 0.1 & 0.212 & 0.134 / 0.444 \\
    & 1 & 0.219 & 0.120 / 0.462 \\
    & 10 & 0.224 & 0.116 / 0.468\\
    \midrule
    \multirow{4}{*}{UTKFace} & 0 & 0.637 & 0.077 / 0.159 \\
    & 0.1 & 0.639 & 0.076 / 0.144 \\
    & 1 & 0.651 & 0.065 / 0.160 \\
    & 10 & 0.651 & 0.058 / 0.148 \\
    \midrule
    \multirow{4}{*}{AdultCensus} & 0 & 0.246 & 0.104 / 0.170\\
    & 0.1 & 0.247 & 0.105 / 0.168 \\
    & 1 & 0.248 & 0.104 / 0.166 \\
    & 10 & 0.248 & 0.104 / 0.165 \\
    \bottomrule
  \end{tabular}
  \caption{Results of \moderate{} when varying $\lambda$.}% (same data amount and budget as Table ~\ref{tbl:systemscollection}).}
  \label{tbl:balancinglambdameasures}
  
  \vspace{-0.8cm}
%   \vspace{-1.0cm}
\end{table}

% \begin{table}[t]
%   \centering
%   \begin{tabular}{c@{\hspace{5pt}}c@{\hspace{5pt}}c@{\hspace{5pt}}c@{\hspace{5pt}}c@{\hspace{5pt}}c@{\hspace{5pt}}c@{\hspace{5pt}}c@{\hspace{5pt}}c@{\hspace{5pt}}c@{\hspace{5pt}}c}
%     \toprule
%     {\bf Method} & {\bf 0} & {\bf 1} & {\bf 2} & {\bf 3} & {\bf 4} & {\bf 5} & {\bf 6} & {\bf 7} & {\bf 8} & {\bf 9} \\
%     \midrule
%     Original & 400 & 400 & 400 & 400 & 400 & 400 & 400 & 400 & 400 & 400 \\
%     $\lambda = 0$ & 734 & 49 & 769 & 495 & 830 & 305 & 1196 & 282 & 278 & 35 \\
%     $\lambda = 0.1$ & 771 & 4 & 848 & 517 & 886 & 245 & 1287 & 218 & 216 & 8\\
%     $\lambda = 1$ & 924 & 0 & 1003 & 366 & 1052 & 58 & 1523 & 33 & 40 & 0\\
%     $\lambda = 10$ & 840 & 0 & 1068 & 380 & 1116 & 0 & 1595 & 0 & 1 & 0\\
%     \bottomrule
%   \end{tabular}
%   \caption{Selective data acquisition results for the Fashion-MNIST experiments in Table~\ref{tbl:balancinglambdameasures}.}
%   \label{tbl:balancinglambdacollection}
% \end{table}

\begin{table}[t]
  \centering
  \begin{tabular}{c@{\hspace{5pt}}c@{\hspace{5pt}}c@{\hspace{5pt}}c@{\hspace{5pt}}c@{\hspace{5pt}}c@{\hspace{5pt}}c@{\hspace{5pt}}c@{\hspace{5pt}}c@{\hspace{5pt}}c@{\hspace{5pt}}c}
    \toprule
    {\bf Method} & {\bf 0} & {\bf 1} & {\bf 2} & {\bf 3} & {\bf 4} & {\bf 5} & {\bf 6} & {\bf 7} & {\bf 8} & {\bf 9} \\
    \midrule
    Original & 200 & 200 & 200 & 200 & 200 & 200 & 200 & 200 & 200 & 200 \\
    $\lambda = 0$ & 755 & 171 & 880 & 606 & 1028 & 312 & 1196 & 402 & 366 & 284\\
    $\lambda = 0.1$ & 921 & 233 & 969 & 504 & 963 & 247 & 1304 & 363 & 199 & 297\\
    $\lambda = 1$ & 915 & 33 & 1301 & 447 & 1451 & 19 & 1677 & 56 & 92 & 9\\
    $\lambda = 10$ & 969 & 0 & 1170 & 277 & 1541 & 0 & 2043 & 0 & 0 & 0\\
    \bottomrule
  \end{tabular}
  \caption{Selective data acquisition results for the Fashion-MNIST experiments in Table~\ref{tbl:balancinglambdameasures}.}
  \label{tbl:balancinglambdacollection}
  \vspace{-0.6cm}
%   \vspace{-0.6cm}
\end{table}

\subsubsection{Balancing $\lambda$}
\label{sec:balancinglambda}

We study the effect of balancing $\lambda$. Recall that a higher $\lambda$ means there is more emphasis on optimizing fairness. How to set $\lambda$ depends on whether the loss or unfairness is more important to minimize for the given application. Table~\ref{tbl:balancinglambdameasures} shows the loss and unfairness results on the four datasets when varying $\lambda$ using the \moderate{} method and the same initial data and budget as in Table~\ref{tbl:systemscollection}. As $\lambda$ increases, the avg.\@ and max.\@ EER results decrease while the loss increases. 

Table~\ref{tbl:balancinglambdacollection} shows the amounts of data acquired per slice for different $\lambda$ values using the Fashion-MNIST dataset. The results for the other datasets (not shown) are similar. In our example, slices \#2, \#4, and \#6 start with higher losses than other slices, and the experiments with higher $\lambda$ values results tend to be more aggressive in acquiring data for those three slices in order to reduce the unfairness.

\begin{table*}[t]
  \centering
%   \color{blue}\begin{tabular}{c@{\hspace{10pt}}c@{\hspace{10pt}}c@{\hspace{10pt}}c@{\hspace{10pt}}c@{\hspace{10pt}}c@{\hspace{10pt}}c@{\hspace{10pt}}c@{\hspace{10pt}}c@{\hspace{10pt}}c}
   \begin{tabular}{c@{\hspace{7pt}}c@{\hspace{7pt}}c@{\hspace{7pt}}c@{\hspace{7pt}}c@{\hspace{7pt}}c@{\hspace{7pt}}c@{\hspace{7pt}}c@{\hspace{7pt}}c@{\hspace{7pt}}c}
  
    \toprule
  {\bf Dataset} & {\bf Measure} & {\bf Alg.} & \multicolumn{2}{c}{\bf Basic} & \multicolumn{2}{c}{\bf Bad for Uniform} & \multicolumn{2}{c}{\bf Bad for Water filling} \\
  
    \toprule
     &  &  & Initial & 3000 & Initial & 3000 & Initial & 3000 \\

    \toprule
    \multirow{6}{*} & \multirow{2}{*}{\em Loss} & {\em Uni} & \multirow{3}{*}{0.472 $\pm$ 0.002} & 0.355 $\pm$ 0.003 & \multirow{3}{*}{0.445 $\pm$ 0.003} & 0.334 $\pm$ 0.002 & \multirow{3}{*}{0.426 $\pm$ 0.002} & 0.322 $\pm$ 0.003 \\
    & \multirow{2}{*}{({\em \# Iters})} & {\em WF} & & 0.355 $\pm$ 0.003 &  & 0.323 $\pm$ 0.002 & & 0.330 $\pm$ 0.003 \\
    {Fashion-}&  & {\em Mod}  &  & {\bf 0.341 $\pm$ 0.002} (3) & &  {\bf 0.314 $\pm$ 0.002} (2) & & {\bf 0.318 $\pm$ 0.002} (3) \\
    \cmidrule{2-9}
    {MNIST}& \multirow{3}{*}{\em Avg. EER} & {\em Uni} & \multirow{3}{*}{0.283 $\pm$ 0.006} & 0.230 $\pm$ 0.004 & \multirow{3}{*}{0.308 $\pm$ 0.009} & 0.226 $\pm$ 0.007 & \multirow{3}{*}{0.210 $\pm$ 0.004 } & 0.177 $\pm$ 0.003 \\
    &  & {\em WF}  &  &  0.230 $\pm$ 0.004 &  & 0.217 $\pm$ 0.005 &  & 0.200 $\pm$ 0.003 \\
    &  & {\em Mod}  & & {\bf 0.173 $\pm$ 0.006} &  & {\bf 0.168 $\pm$ 0.003} &  & {\bf 0.157 $\pm$ 0.002 } \\

    \midrule
    %mixed-mnist
    \multirow{6}{*} & \multirow{2}{*}{\em Loss} & {\em Uni} & \multirow{3}{*}{0.301 $\pm$ 0.003} & 0.240 $\pm$ 0.002 & \multirow{3}{*}{0.264 $\pm$ 0.003 } & 0.229 $\pm$ 0.002 & \multirow{3}{*}{0.266 $\pm$ 0.002} & 0.226 $\pm$ 0.002\\
    & \multirow{2}{*}{({\em \# Iters})} & {\em WF} & & 0.240 $\pm$ 0.002 &  & 0.221 $\pm$ 0.002 & & 0.228 $\pm$ 0.001 \\
    {Mixed-}&  & {\em Mod}  &  & {\bf 0.235 $\pm$ 0.001 } (3) & &  {\bf 0.216 $\pm$ 0.002} (2) & & {\bf 0.219 $\pm$ 0.002} (3) \\
    \cmidrule{2-9}
    {MNIST}& \multirow{3}{*}{\em Avg. EER} & {\em Uni} & \multirow{3}{*}{0.207 $\pm$ 0.004 } & 0.182 $\pm$ 0.002 & \multirow{3}{*}{0.205 $\pm$ 0.004} & 0.182 $\pm$ 0.002 & \multirow{3}{*}{0.166 $\pm$ 0.003} & 0.157 $\pm$ 0.003 \\
    &  & {\em WF}  &  &  0.182 $\pm$ 0.002 &  & 0.166 $\pm$ 0.002 &  & 0.167 $\pm$ 0.002 \\
    &  & {\em Mod}  & & {\bf 0.147 $\pm$ 0.002} &  & {\bf 0.146 $\pm$ 0.002 } &  & {\bf 0.133 $\pm$ 0.003 } \\
    \midrule

    %utkface
    \multirow{6}{*}{UTKFace} & \multirow{2}{*}{\em Loss} & {\em Uni} & \multirow{3}{*}{0.562 $\pm$ 0.003} & 0.554 $\pm$ 0.005 & \multirow{3}{*}{0.611 $\pm$ 0.005} & 0.579 $\pm$ 0.003 & \multirow{3}{*}{0.597 $\pm$ 0.004} & 0.575 $\pm$ 0.004 \\
    & \multirow{2}{*}{({\em \# Iters})} & {\em WF} & & 0.554 $\pm$ 0.005 &  & 0.573 $\pm$ 0.004 & & 0.577 $\pm$ 0.003\\
    &  & {\em Mod}  &  & {\bf 0.547 $\pm$ 0.004} (1) & &  {\bf 0.571 $\pm$ 0.004} (1) & & {\bf 0.569 $\pm$ 0.003} (1) \\
    \cmidrule{2-9}
    & \multirow{3}{*}{\em Avg. EER} & {\em Uni} & \multirow{3}{*}{0.091 $\pm$ 0.005} & 0.081 $\pm$ 0.009 & \multirow{3}{*}{0.117 $\pm$ 0.006} & 0.095 $\pm$ 0.004 & \multirow{3}{*}{0.093 $\pm$ 0.007} & 0.087 $\pm$ 0.006 \\
    &  & {\em WF}  &  &  0.081 $\pm$ 0.009 &  & 0.090 $\pm$ 0.004 &  & 0.074 $\pm$ 0.005 \\
    &  & {\em Mod}  & & {\bf 0.071 $\pm$ 0.005} &  & {\bf 0.078 $\pm$ 0.005} &  & {\bf 0.072 $\pm$ 0.004} \\
    
    \toprule
    &  &  & Initial & 300 & Initial & 300 & Initial & 300 \\
    \toprule
    %adult
    \multirow{6}{*}{AdultCensus} & \multirow{2}{*}{\em Loss} & {\em Uni} & \multirow{3}{*}{0.263 $\pm$ 0.001} & 0.257 $\pm$ 0.001 & \multirow{3}{*}{0.258 $\pm$ 0.001} & 0.254 $\pm$ 0.001& \multirow{3}{*}{0.287 $\pm$ 0.001} & 0.286 $\pm$ 0.001 \\
    & \multirow{2}{*}{({\em \# Iters})} & {\em WF} & & 0.257 $\pm$ 0.001 &  & 0.247 $\pm$ 0.000 & & 0.288 $\pm$ 0.001\\
    &  & {\em Mod}  &  & {\bf 0.253 $\pm$ 0.001} (1) & &  {\bf 0.245 $\pm$ 0.000} (1) & & {\bf 0.284 $\pm$ 0.001} (2) \\

    \cmidrule{2-9}
    & \multirow{3}{*}{\em Avg. EER} & {\em Uni} & \multirow{3}{*}{0.103 $\pm$ 0.001} & 0.099 $\pm$ 0.000 & \multirow{3}{*}{0.109 $\pm$ 0.001} & 0.106 $\pm$ 0.000 & \multirow{3}{*}{0.137 $\pm$ 0.001} & 0.137 $\pm$ 0.002\\
    &  & {\em WF}  &  &  0.099 $\pm$ 0.000 &  & 0.101 $\pm$ 0.000 &  & 0.137 $\pm$ 0.001\\
    &  & {\em Mod}  & & {\bf 0.094 $\pm$ 0.000} &  & {\bf 0.100 $\pm$ 0.000} &  & {\bf 0.134 $\pm$ 0.001} \\

    \bottomrule
  \end{tabular}
  \caption{A detailed comparison of \moderate{}{\em (Mod)} against the two baselines -- {\em Uniform(Uni)} and {\em Water filling(WF)} -- on the four datasets where we use three settings and vary the budget $B$. We also set $\lambda = 0.1$.}
  \label{tbl:baselinecomparisons}
  \vspace{-0.5cm}
\end{table*}

\subsubsection{Comparison with Baselines}
\label{sec:comparisonbaselines}
% use tables instead of figures?
We make a detailed comparison between \moderate{} and the baselines {\em Uniform} and {\em Water filling} in Table~\ref{tbl:baselinecomparisons} where we use three settings: (1) a basic setting where slices have the same amounts of data, (2) a pathological setting for {\em Uniform} (called ``Bad for Uniform'') where there are many slices with low loss, and (3) a pathological setting for {\em Water filling} (called ``Bad for Water filling'') where there is a large slice with high loss and a small slice with low loss. For each setting, we compare the three methods using a budget of $B$ = 3K except for the AdultCensus dataset where $B$ = 300 is already enough to obtain low loss. As a result, \moderate{} has both lower loss and avg.\@ EER than the baselines for all datasets. Looking at the two baselines, {\em Uniform} performs worse than {\em Water filling} in setting (2) while {\em Water filling} performs worse in setting (3). For all experiments, \moderate{} mostly performs 1--3 iterations as shown in the parentheses in Table~\ref{tbl:baselinecomparisons}. 

\begin{figure}[t]
\centering
  \includegraphics[width=\columnwidth]{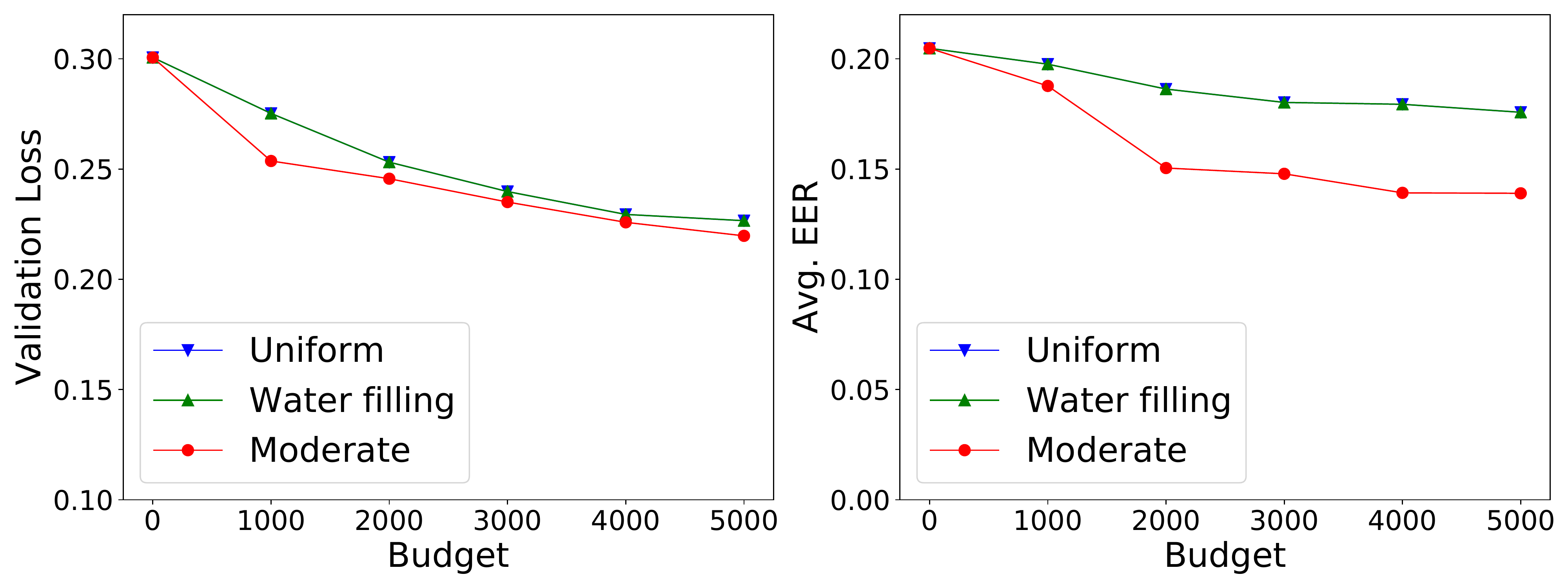}
  \vspace{-0.8cm}
  \caption{Loss and unfairness comparison (Mixed-MNIST).}
  \label{fig:baselinecomparisons}
%   \vspace{-0.2cm}
  \vspace{-0.4cm}
\end{figure}

Figure~\ref{fig:baselinecomparisons} shows a more detailed comparison with the baselines using the basic setting above (i.e., setting (1)) and the Mixed-MNIST dataset. Here the two baselines happen to have the same loss and unfairness results. For varying budgets, \moderate{} clearly outperforms the two baselines in terms of loss and unfairness. Although the improvements in loss are not as drastic as those in unfairness, they are still significant: in order to obtain the same losses, the baselines need to increase their budgets by 15--100\%.

\kh{We also perform the experiments with ResNet-18~\cite{resnet} (one of the state-of-the-art models for image classification) using the basic setting and the Fashion-MNIST dataset. The results are in Appendix ~\ref{sec:resnet}, and the trends are similar to Table~\ref{tbl:baselinecomparisons} where \moderate{} outperforms the baselines in terms of loss and unfairness.}

% The results are in our technical report~\cite{slicetunertr}, and the trends are similar to Table~\ref{tbl:baselinecomparisons} where \moderate{} outperforms the baselines in terms of loss and unfairness.

%In fact, Baidu~\cite{baidu2017deep} empirically validates a power-law scaling relationship for state-of-the-art model architectures in four machine learning domains, which demonstrates that \systems{} can be applied not only to simple models but also to state-of-the-art models.}

\subsubsection{Unreliable learning curves}
\label{sec:unreliablelearningcurves}

%\subsubsection{Small Data Results}
%\label{sec:smalldata}

Learning curves are not always ``perfect'' in the sense that they cleanly follow a power law. For example, if the slice size is small, then the learning curve is more likely to be noisy. Even if the slice size is not small, some slices may still exhibit noisy learning curves for other reasons.

\begin{figure}[t]
\centering
  \includegraphics[width=\columnwidth]{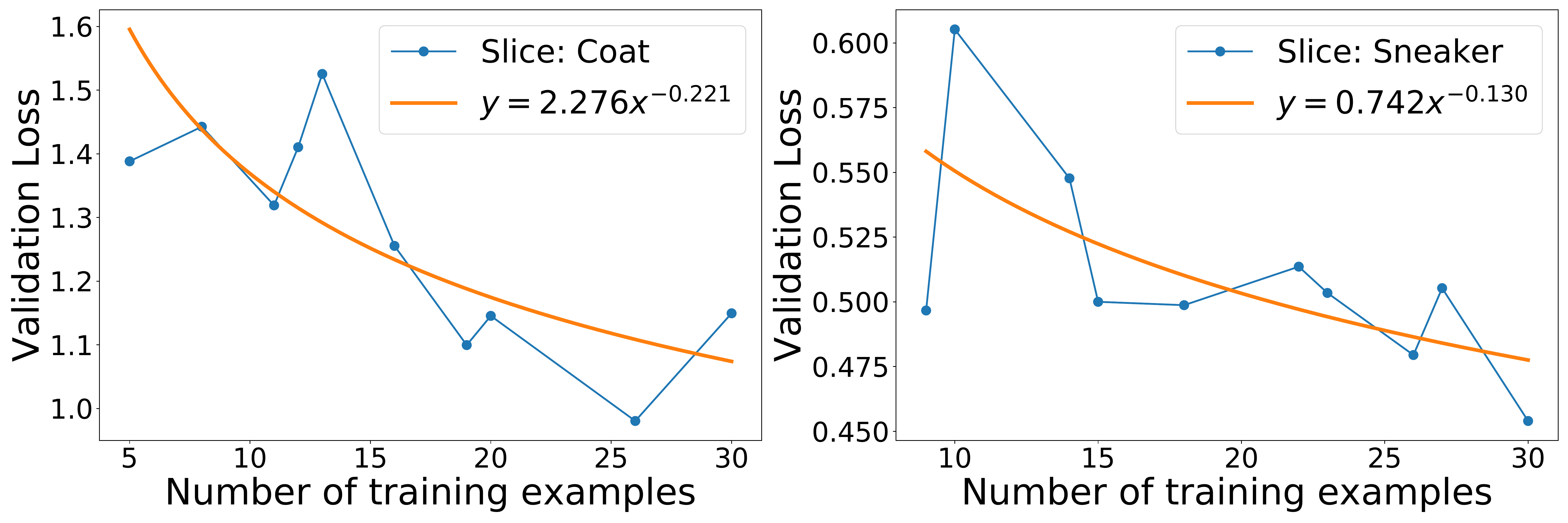}
  \vspace{-0.8cm}
  \caption{Learning curves are noisy for small slices of Fashion-MNIST.}
  \label{fig:smalldata}
%   \vspace{-0.2cm}
  \vspace{-0.6cm}
\end{figure}

In this section, we consider the scenario of small slices where we lower the initial slice sizes of the Fashion-MNIST dataset to $L = 30$ where the learning curves are noisy and unreliable as shown in Figure~\ref{fig:smalldata}. Nonetheless, Table~\ref{tbl:smalldata} shows that \systems{} still performs better than the baselines because it can leverage the relative differences (loss and steepness) of the learning curves. Even if the learning curves are completely useless, \systems{}'s performance should fall back to those of the baselines. In a sense, the learning curves are used as hints and do not have to be perfect. However, as more data is acquired, the learning curves will gradually become more reliable and informative.

\begin{table}[t]
  \centering
  \begin{tabular}{cccc}
    \toprule
  {\bf Init. Size} & {\bf Method} & {\bf Loss} & {\bf Avg. / Max. EER} \\
    \midrule
    \multirow{4}{*}{} & Original & 0.583 & 0.317 / 0.792 \\
    {30} & Uniform & 0.508 & 0.302 / 0.786 \\
    {($B = 500$)}& Water filling & 0.508 & 0.302 / 0.786 \\
    & \moderate{} & {\bf 0.475} & {\bf 0.259} / {\bf 0.650} \\
    \bottomrule
  \end{tabular}
  \caption{Loss and unfairness results of \systems{} methods for the small slice sizes of Fashion-MNIST.}
  \label{tbl:smalldata}
  \vspace{-0.6cm}
\end{table}

\subsection{Efficient Learning Curve Generation}
\label{sec:efficientlearningcurvegeneration}

We evaluate the efficient learning curve generation method described in Section~\ref{sec:efficientimplementation} as shown in Table~\ref{tbl:efficientlearningcurves}. The default \moderate{} method uses the efficient learning curve generation. We make a comparison with a modified version of \moderate{} where the learning curve generation is done exhaustively (called ``Exhaustive''). We also vary the initial slice size and budget. As a result, \moderate{} is 11-12x faster than Exhaustive as expected and \kh{obtains slightly worse or even better loss and unfairness results. While it may sound counter intuitive that \moderate{} can have lower loss and unfairness, this may happen because our optimization of taking X\% examples of all slices together has the effect of removing bias among the slices.}
% similar loss and unfairness results. 
The reason the runtime speedups are larger than the number of slices (10) is that the model training itself is on average faster because it is performed on smaller data as explained in Section~\ref{sec:efficientimplementation}. We consider the runtimes of \moderate{} to be practical because the main bottleneck of \systems{} is the time to actually acquire data, which may take say hours.

% \begin{table}[t]
%   \centering
%   \begin{tabular}{c@{\hspace{5pt}}c@{\hspace{5pt}}c@{\hspace{5pt}}c@{\hspace{5pt}}c}
%     \toprule
%     {\bf Method} & {\bf Loss} & {\bf Avg. / Max. EER} & {\bf Runtime (sec)} \\
%     \midrule
%     \multicolumn{4}{c}{Init. size = 200, $B$ = 2K}\\
%     \midrule
%     Exhaustive & 0.613 & 0.275 / 0.479 & 49,211 \\
%     \moderate{} & 0.610 & 0.239 / 0.624 & 891 \\
%     \midrule
%     \multicolumn{4}{c}{Init. size = 300, $B$ = 3K}\\
%     \midrule
%     Exhaustive & 0.546 & 0.245 / 0.424 & 40,387 \\
%     \moderate{} & 0.540 & 0.247 / 0.467 & 1,244 \\
%     \bottomrule
%   \end{tabular}
%   \caption{A comparison between a method using exhaustive generation of learning curves versus the \moderate{} method on the Fashion-MNIST dataset for different initial slice sizes and budget values. }
%   \label{tbl:efficientlearningcurves}
% \end{table}

\begin{table}[t]
  \centering
  \begin{tabular}{c@{\hspace{5pt}}c@{\hspace{5pt}}c@{\hspace{5pt}}c@{\hspace{5pt}}c}
    \toprule
    {\bf Method} & {\bf Loss} & {\bf Avg. / Max. EER} & {\bf Runtime (sec)} \\
    \midrule
    \multicolumn{4}{c}{Init. size = 200, $B$ = 2K}\\
    \midrule
    Exhaustive & 0.351 & 0.159 / 0.388 & 20,275 \\
    \systems{} & 0.346 & 0.164 / 0.362 & 1,847 \\
    \midrule
    \multicolumn{4}{c}{Init. size = 300, $B$ = 3K}\\
    \midrule
    Exhaustive & 0.301 & 0.138 / 0.375 & 24,673 \\
    \systems{} & 0.303 & 0.137 / 0.303 & 2,088 \\
    \bottomrule
  \end{tabular}
  \caption{A comparison between exhaustively generating learning curves versus the \moderate{} method on Fashion-MNIST for different initial slice sizes and budget values. }
  \label{tbl:efficientlearningcurves}
  \vspace{-0.8cm}
%   \vspace{-1.5cm}
\end{table}

\section{Related Work}
\label{sec:relatedwork}

\paragraph*{Data Acquisition}

Acquiring data is becoming increasingly convenient. Dataset searching~\cite{DBLP:journals/pvldb/NargesianZMPA19} focuses on finding the right datasets either on the Web or within a data lake. As a recent example, Google Dataset Search~\cite{DBLP:journals/corr/abs-2006-06894} started its service in 2016 with 500K public science datasets, but now has 30M datasets. Another traditional approach is crowdsourcing~\cite{amazonmechanicalturk}, where one can hire workers to generate new data. Yet another interesting branch of research is to use simulators~\cite{DBLP:conf/aaai/KimLHS19} to generate infinite numbers of realistic examples. \systems{} complements these systems by determining how much data to acquire per slice based on a given ML model.

Although the notion of selective data acquisition seems relevant to active learning~\cite{DBLP:series/synthesis/2012Settles}, they solve different problems. While active learning selects which existing unlabeled data to label, \systems{} decides how much non-existing data to acquire per slice. Hence, the active learning techniques do not apply for data acquisition.

%\kh{Recently, reference ~\cite{abernethy2020adaptive} proposes a fair data acquisition algorithm, which adaptively samples a data point from the group that is currently disadvantaged. Compared to \systems{}, their strategy only focuses on the most disadvantaged group, although more acquisition has diminishing returns. Moreover, their approach acquires one data point at each iteration.}

% While \systems{} selectively acquires data by estimating the amount of error reduction for all slices through the learning curves

The closest work to \systems{} is reference~\cite{DBLP:conf/nips/ChenJS18}, which also acquires data to improve model accuracy and fairness. A key observation is that unfairness can be decomposed into bias, variance, and noise, and that data acquisition can reduce unfairness without sacrificing accuracy. Similar to \systems{}, reference~\cite{DBLP:conf/nips/ChenJS18} assumes that learning curves follow a power law. However, the learning curves are primarily used to improve fairness, and the data acquisition method is simplistic where the amounts of data acquired follow the original data distribution. In comparison, \systems{} takes the more sophisticated approach of using learning curves to optimize both accuracy and fairness and addresses the challenge of unreliable learning curves using efficient iterative updates.

%\kh{The goal of reference ~\cite{Asudeh2019AssessingAR} is to acquire data such that all possible slices (=regions of attribute space) have enough size (=coverage). }

\kh{Another close work~\cite{Asudeh2019AssessingAR} performs data acquisition such that all possible slices are large enough. This problem is analogous to Water filling (Figure~\ref{fig:baseline2}) where the slices are guaranteed a minimal data coverage, but in the more challenging setting where slices may overlap with each other, and there is a combinatorial explosion in the number of slices. In comparison, \systems{} assumes the slices partition the entire dataset and selectively acquires data to improve accuracy and fairness instead of achieving minimum coverage.}

\paragraph*{\kh{Multi-armed Bandit}}
%\kh{The multi-armed bandit~\cite{10.1023/A:1013689704352} is a well-studied problem to model the exploration-exploitation dilemma.}
\kh{\systems{} can be viewed as solving a specialized multi-armed bandit problem~\cite{10.1023/A:1013689704352}. Among the known variants, rotting bandit (RB) ~\cite{rottingbandit} is the most relevant where each arm's expected reward decays as a function of the pulls of that arm, which corresponds to the diminishing benefit of newly-acquired data in our setting. However, there is no fairness notion of ensuring the rewards across slots are similar in RB. Moreover, RB only uses prior knowledge on the rotting models, but we use prior knowledge on the reward models including rotting, which we capture as power-law learning curves. Our prior knowledge is not strong because learning curves may change as we acquire data, but can be used to select multiple slots that together minimize loss and unfairness. Finally, RB assumes independent arms while we assume dependence where pulling one arm may change the expected rewards of others.}

\paragraph*{Model Fairness}

As ML is used more broadly, model fairness is becoming a critical issue~\cite{DBLP:conf/pods/Venkatasubramanian19}.
Fairness is a subjective notion, so there is no universally-agreed definition.
Some standard definitions include demographic parity~\cite{DBLP:conf/kdd/FeldmanFMSV15}, equalized odds~\cite{DBLP:conf/nips/HardtPNS16}, equal opportunity~\cite{DBLP:conf/nips/HardtPNS16}, and equalized error rates~\cite{DBLP:conf/pods/Venkatasubramanian19,DBLP:conf/www/ZafarVGG17}.
Other popular notions of fairness include individual fairness~\cite{Dwork:2012:FTA:2090236.2090255} and causality-based fairness~\cite{10.5555/3294771.3294834}.
Among them, we choose equalized error rates because it is relevant to ML systems that must provide similar-quality services to users. 
In addition, equalized error rates is a familiar concept in the systems literature where the system performance must be uniform across data partitions.

% \kh{More recently, there has been a surge in unfairness mitigation techniques~\cite{DBLP:journals/corr/abs-1810-01943,DBLP:conf/bigdataconf/XuYZW18}, which improve model fairness by either fixing the training data (pre-processing), model training (in-processing), or the trained model (post-processing). Most of these approaches assume a fixed dataset, and \systems{} complements them by selectively acquiring new data for different slices to improve model fairness. }

\paragraph*{Model Analysis and Learning Curves}

An ML model should be viewed as software and thus needs to be analyzed and debugged as well.
Industrial tools like TensorFlow Model Analysis~\cite{tfma} provide visualizations of model accuracy per slice. 
More recently, automatic data slicing techniques~\cite{slicefinder} have been proposed to find underperforming slices. Estimating learning curves has been mainly studied in the computer vision community~\cite{baidu2017deep,DBLP:journals/corr/ChoLSCD15,DBLP:journals/midm/FigueroaZKN12,DBLP:journals/jbi/Hajian-Tilaki14}.
Baidu~\cite{baidu2017deep} analyzes four ML domains of how the learning curve changes depending on the amount of data and observes a power-law error scaling. This result is further supported theoretically~\cite{AMARI1993161} and empirically~\cite{DBLP:conf/ijcai/DomhanSH15}. Reference~\cite{DBLP:conf/nips/ChenJS18} also utilizes a power-law learning curve to acquire data and improve fairness.

\section{Conclusion}

We proposed \systems{}, which performs selective data acquisition to optimize model accuracy and fairness. \systems{} estimates learning curves for different slices and solving an optimization problem to determine how much data acquisition is needed for each slice given a budget. To address the challenges of unreliable learning curves and dependencies among slices, we proposed iterative algorithms that repeatedly update the learning curves as more data is acquired, using the imbalance ratio change as a proxy to estimate influence. We demonstrated on real datasets that our iterative algorithms are efficient and obtain lower loss and unfairness than the two baseline methods that do not exploit the learning curves, even if the learning curves are unreliable. In addition, we demonstrated the practicality of \systems{} by acquiring new data using Amazon Mechanical Turk. \kh{In the future, we would like to improve our influence estimation and support overlapping slices.}

%better formalize the notion of influence among slices based on their contents and extend our techniques to support overlapping slices.}

\paragraph*{Acknowledgments}

This work was supported by a Google AI Focused Research Award and by the National Research Foundation of Korea(NRF) grant funded by the Korea government(MSIT) (No. NRF-2018R1A5A1059921 and NRF-2021R1C1C1005999).

\balance
\bibliographystyle{ACM-Reference-Format}
\bibliography{main}

\clearpage 
\newpage

\appendix

\section{Data slicing}
\label{sec:slicingmethods}

In this section, we briefly discuss various approaches for data slicing. While this topic is not the main focus of this paper, it is worth discussing the state-of-the-art methods. The straightforward way is to select slices manually based on domain knowledge. For example, for a movie recommendation system, one may select slices based on genre. Alternatively, one can determine slices based on model analysis. Manual tools for visualization~\cite{tfma, kahng2016visual} can be used to find problematic slices where a model underperforms. Recently, there are automatic tools~\cite{slicefinder} that can find such slices.

As we mentioned in Section~\ref{sec:preliminaries}, a desirable property of a slice is to be unbiased so that the acquired examples have similar effects on the model accuracy. Using large slices that are biased is undesirable. For example, a slice that contains all regions of Figure~\ref{fig:customers1} is bad because there is a bias towards American customers. That is, adding an American customer example is not as helpful as say adding a European customer example. On the other hand, using slices that are not biased, but too small is also problematic because we may need to maintain many learning curves that are not accurate due to the small amounts of data. 

In order to find the largest-possible slices that are still unbiased, one can use a method similar to decision tree training where the goal is to find partitions of the data such that the impurity (i.e., homogeneity of labels in leaf nodes) is minimized. Instead of minimizing impurity, we would have to compute the bias in slices using say an entropy-based measure. Starting from the entire dataset, we can iteratively split slices that have biases in their data for different values of attributes. The splitting can terminate once the average entropy is above some threshold.

\section{ResNet-18 Results on Fashion-MNIST}
\label{sec:resnet}
% In this section, we perform the experiments on ResNet-18~\cite{resnet} (one of the state-of-the-art models in an image classification task) using the basic setting and Fashion-MNIST dataset. In Table~\ref{tbl:sotamodel} below, the key trends are similar to Table 6 where \moderate{} outperforms baselines in terms of loss and fairness. In fact, Baidu~\cite{baidu2017deep} empirically validates a power-law scaling relationship for state-of-the-art model architectures in four machine learning domains, which demonstrates that \systems{} can be applied not only to simple models but also to state-of-the-art models.

In Section ~\ref{sec:experiments}, we used basic convolution neural networks and single fully connected layers for all experiments. In this section, we performed the same experiments with ResNet-18~\cite{resnet} (one of the state-of-the-art models for image classification) using the basic setting and Fashion-MNIST dataset. In Table~\ref{tbl:sotamodel}, the key trends are still similar to Table~\ref{tbl:baselinecomparisons} where Moderate outperforms baselines in terms of loss and fairness. Compared to the Fashion-MNIST results in Table~\ref{tbl:baselinecomparisons}, the losses actually increase because ResNet-18's architecture is overly complex to train on the modest-sized Fashion-MNIST dataset.

% We agree with your point on improving on state-of-the-art methods and performed new experiments with ResNet-18~\cite{resnet} (one of the state-of-the-art models for image classification) using the basic setting and Fashion-MNIST dataset. In Table~\ref{tbl:sotamodel} below, the key trends are similar to Table 6 in the paper where \moderate{} still outperforms the baselines in terms of loss and fairness. Compared to the Fashion-MNIST results in Table 6, the losses actually increases because ResNet-18's architecture is overly complex to train on the modest-sized Fashion-MNIST dataset. We added Table~\ref{tbl:sotamodel} in our technical report and this discussion in \kh{the last paragraph of Section 6.3.3}. We also uploaded the relevant source code and jupyter results to the Fashion-MNIST folder in our GitHub~\cite{github}.

\begin{table}[ht]
  \centering
  \begin{tabular}{cccc}
    \toprule
  {\bf Init. Size} & {\bf Method} & {\bf Loss} & {\bf Avg. / Max. EER} \\
    \midrule
    \multirow{4}{*}{} & Original & 0.567 & 0.240 / 0.481 \\
    {400} & Uniform & 0.484 & 0.228 / 0.452 \\
    {($B = 3K$)}& Water filling & 0.484 & 0.228 / 0.452 \\
    & \moderate{} & {\bf 0.480} & {\bf 0.190} / {\bf 0.422} \\
    \bottomrule
  \end{tabular}
  \caption{Loss and unfairness results of \systems{} methods for the ResNet-18 on Fashion-MNIST.}
  \label{tbl:sotamodel}
\end{table}

\section{Exponential Distribution Results}
\label{sec:exponential}

In Section~\ref{sec:lossunfairness}, we evaluated the \systems{} methods when the initial slice sizes are the same. In this section, we perform the same experiments when the initial slice sizes follow an exponential distribution. In Table~\ref{tbl:systemsmeasures_zipf}, the key trends are similar to Table~\ref{tbl:systemsmeasures} where iterative algorithms outperform \oneshot{} because \oneshot{} tends to acquire too much data for certain slices as shown in Table~\ref{tbl:systemscollection_zipf}. Moreover, while \linear{} uses more iterations than \exponential{} and \moderate{}, it has slightly-better loss and unfairness results.

% In Figure~\ref{tbl:systemscollection_zipf}, the amounts of data acquired for Moderate are identical to Aggressive. We anticipate that for larger budgets, Moderate will have different performance than Aggressive. 

\begin{table}[htbp]
  \centering
  \begin{tabular}{cccc}
    \toprule
    {\bf Dataset} & {\bf Method} & {\bf Loss} & {\bf Avg./Max. EER} \\ %& {\bf \# Iters} \\
    \midrule
    \multirow{4}{*}{\begin{tabular}{@{}c@{}}Fashion- \\ MNIST\end{tabular}} & Original & 0.455 & 0.260 / 0.911 \\ %& n/a \\
    {} & \oneshot{} & 0.349 & 0.178 / 0.448 \\ %& 1 \\
    {} & \exponential{} & 0.312 & {\bf 0.116} / {\bf 0.232} \\ %& 3 \\
    {} & \moderate{} & 0.308 & 0.125 / 0.306 \\ %& 3 \\
    & \linear{} & {\bf 0.307} & {\bf 0.116} / 0.299 \\ %& 5 \\
    \midrule
    \multirow{4}{*}{\begin{tabular}{@{}c@{}}Mixed- \\ MNIST\end{tabular}} & Original & 0.292 & 0.200 / 0.903 \\ %& n/a \\
    {} & \oneshot{} & 0.249 & 0.157 / 0.934 \\ %& 1 \\
    {} & \exponential{} & 0.231 & 0.124 / 0.465 \\ %& 3 \\
    {} & \moderate{} & 0.231 & 0.124 / 0.465 \\ %& 3 \\
    & \linear{} & {\bf 0.223} & {\bf 0.117} / {\bf 0.381} \\ %& 5 \\
    \midrule
    \multirow{4}{*}{UTKFace} & Original & 0.710 & 0.170 / 0.369 \\ %& n/a \\
    & \oneshot{} & 0.655 & 0.114 / 0.285 \\ %& 1 \\
    & \exponential{} & {\bf 0.638} & {\bf 0.114} / {\bf 0.285} \\ %& 2 \\
    & \moderate{} & {\bf 0.638} & {\bf 0.114} / {\bf 0.285} \\ %& 2 \\
    & \linear{} & {\bf 0.638} & {\bf 0.114} / {\bf 0.285} \\ %& 3 \\
    \midrule
    \multirow{4}{*}{AdultCensus} & Original & 0.274 & 0.107 / 0.172 \\ %& n/a \\
    & \oneshot{} & 0.253 & {\bf 0.095} / 0.150 \\ %& 1 \\
    & \exponential{} & {\bf 0.252} & {\bf 0.095} / {\bf 0.149} \\ %& 2 \\
    & \moderate{} & {\bf 0.252} & {\bf 0.095} / {\bf 0.149} \\ %& 2 \\
    & \linear{} & {\bf 0.252} & {\bf 0.095} / {\bf 0.149} \\ %& 3 \\
    \bottomrule
  \end{tabular}
  \caption{\systems{} methods comparison on the 4 datasets where the initial slice sizes follow an exponential distribution as shown in Table~\ref{tbl:systemscollection_zipf}.}
  \label{tbl:systemsmeasures_zipf}
  %\vspace{-0.4cm}
\end{table}

\begin{table*}[thbp]
  \centering
  \begin{tabular}{ccccccccccccc}
    \toprule
  {\bf Dataset} & {\bf Method} & {\bf 0} & {\bf 1} & {\bf 2} & {\bf 3} & {\bf 4} & {\bf 5} & {\bf 6} & {\bf 7} & {\bf 8} & {\bf 9} & {\bf \# iterations} \\
    \midrule
    \multirow{4}{*}{\begin{tabular}{@{}c@{}}Fashion-MNIST \\ ($B = 6K$)\end{tabular}} & Original & 400 & 282 & 230 & 200 & 178 & 163 & 151 & 141 & 133 & 126 & n/a \\
    & \oneshot{} & 0 & 0 & 596 & 192 & 1495 & 201 & 3019 & 371 & 128 & 0 & 1 \\
    & \exponential{} & 434 & 24 & 1406 & 348 & 1299 & 77 & 2136 & 88 & 188 & 0 & 4 \\
    & \moderate{} & 373 & 74 & 1602 & 1737 & 396 & 617 & 61 & 558 & 965 & 42 & 5 \\
    & \linear{} & 727 & 51 & 1666 & 1604 & 641 & 780 & 47 & 951 & 790 & 79 & 9 \\
    \midrule
    \multirow{4}{*}{\begin{tabular}{@{}c@{}}Mixed-MNIST \\ ($B = 6K$)\end{tabular}} & Original & 600 & 346 & 268 & 200 & 189 & 180 & 166 & 160 & 154 & 145 & n/a \\
    & \oneshot{} & 0 & 0 & 0 & 176 & 61 & 0 & 1457 & 673 & 1136 & 2008 & 1 \\
    & \exponential{} & 0 & 0 & 0 & 57 & 32 & 1432 & 946 & 653 & 1309 & 1356 & 3\\
    & \moderate{} & 0 & 0 & 0 & 57 & 32 & 1432 & 946 & 653 & 1309 & 1356 & 3\\
    & \linear{} & 0 & 0 & 0 & 90 & 59 & 800 & 936 & 730 & 1240 & 1562 & 6 \\
    \midrule
    \multirow{4}{*}{\begin{tabular}{@{}c@{}}UTKFace \\ ($B = 3K$)\end{tabular}} & Original & 400 & 263 & 206 & 174 & 152 & 136 & 124 & 114 & - & - & n/a \\
    & \oneshot{} & 23 & 123 & 245 & 315 & 322 & 279 & 745 & 276 & - & - & 1 \\
    & \exponential{} & 90 & 138 & 146 & 308 & 367 & 296 & 574 & 378 & - & - & 2 \\
    & \moderate{} & 90 & 138 & 146 & 308 & 367 & 296 & 574 & 378 & - & - & 2 \\
    & \linear{} & 90 & 138 & 146 & 308 & 367 & 296 & 574 & 378 & - & - & 2 \\
    \midrule
    \multirow{4}{*}{\begin{tabular}{@{}c@{}}AdultCensus \\ ($B = 500$)\end{tabular}} & Original & 150 & 106 & 86 & 75 & - & - & - & - & - & - & n/a \\
    & \oneshot{} & 0 & 0 & 425 & 75 & - & - & - & - & - & - & 1\\
    & \exponential{} & 0 & 0 & 232 & 268 & - & - & - & - & - & - & 2\\
    & \moderate{} & 0 & 0 & 232 & 268 & - & - & - & - & - & - & 2\\
    & \linear{} & 0 & 0 & 232 & 268 & - & - & - & - & - & - & 2\\
    \bottomrule
  \end{tabular}
  \caption{Selective data acquisition results for the experiments in Table~\ref{tbl:systemsmeasures_zipf} where slices are listed numbered from 0 up to 9 (the ordering is not important), and their initial sizes follow an exponential distribution. For Mixed-MINST, we select 10 out of 20 slices. The {\em Original} row contains the initial number of examples for each slice while the other rows show the additional number of examples acquired.}
  \label{tbl:systemscollection_zipf}
  %\vspace{-0.1cm}
\end{table*}

\end{document}